\documentclass{article}

\PassOptionsToPackage{numbers, compress}{natbib}

\usepackage[final]{neurips_2025}

\usepackage[utf8]{inputenc} 
\usepackage[T1]{fontenc}    
\usepackage{hyperref}       
\usepackage{url}            
\usepackage{amsfonts}       
\usepackage{nicefrac}       
\usepackage{xcolor}         

\usepackage{microtype}
\usepackage{graphicx}
\usepackage{subfigure}
\usepackage{booktabs} 

\usepackage{amsmath}
\usepackage{amssymb}
\usepackage{mathtools}
\usepackage{amsthm}

\usepackage[capitalize,noabbrev]{cleveref}

\usepackage{bbm}
\usepackage{xfrac}
\usepackage{physics}
\usepackage{mdframed}
\usepackage{wrapfig}
\usepackage{bm}
\usepackage[perpage]{footmisc}

\theoremstyle{plain}
\newtheorem{theorem}{Theorem}[section]

\theoremstyle{definition}
\newtheorem{definition}[theorem]{Definition}

\theoremstyle{remark}

\usepackage[textsize=tiny]{todonotes}
\usepackage{framed}

\newcommand{\expect}{{\mathbb{E}}} \newcommand{\bX}{{\boldsymbol{X}}}
\newcommand{\bY}{{\boldsymbol{Y}}} \newcommand{\bZ}{{\boldsymbol{Z}}}
\newcommand{\bW}{{\boldsymbol{W}}} \newcommand{\bx}{{\boldsymbol{x}}}
\newcommand{\by}{{\boldsymbol{y}}} \newcommand{\bz}{{\boldsymbol{z}}}
 \newcommand{\boldf}{{\boldsymbol{f}}}
\newcommand{\bQ}{{\boldsymbol{Q}}} \newcommand{\LsquaredM}{{L$^2$M}}
\newcommand{\btheta}{{\bm{\theta}}} \newcommand{\PG}{\textsc{PG19}}
\newcommand{\Wikipedia}{\textsc{Wikipedia}} \newcommand{\Books}{\textsc{Books3}}

\title{\LsquaredM: Mutual Information Scaling Law for Long-Context Language
Modeling}

\author{%
  Zhuo Chen$^{12}$\thanks{Corresponding author} \hspace{2em} Oriol Mayné i
  Comas$^{23}$ \hspace{2em} Zhuotao Jin$^{24}$ \\
  \textbf{Di Luo$^{1245}$ \hspace{2em} Marin Soljačić$^{12}$} \\
  $^1$ NSF AI Institute for Artificial Intelligence and Fundamental Interactions
  \\
  $^2$ Massachusetts Institute of Technology \\
  $^3$ Polytechnic University of Catalonia \\
  $^4$ Harvard University \\
  $^5$ University of California, Los Angeles \\
  \texttt{{\{\href{mailto:chenzhuo@mit.edu}{chenzhuo},
\href{mailto:omayne@mit.edu}{omayne}, \href{mailto:jinzhta@mit.edu}{jinzhta},
\href{mailto:diluo@mit.edu}{diluo},
\href{mailto:soljacic@mit.edu}{soljacic}\}@mit.edu}} }

\begin{document}

\maketitle

\begin{abstract}

We present a universal\footnote[2]{Our framework applies to autoregressive
language models, which encompass all widely-used LLMs such as GPT, Claude,
Gemini, and LLaMA.} theoretical framework for understanding \emph{long-context
language modeling} based on a \emph{bipartite} mutual information scaling law
that we rigorously verify in natural language. 
We demonstrate that bipartite mutual information captures multi-token
interactions distinct from and scaling independently of conventional two-point
mutual information, and show that this provides a more complete characterization
of the dependencies needed for accurately modeling long sequences.
Leveraging this scaling law, we formulate the {\bf L}ong-context {\bf L}anguage
{\bf M}odeling ({\bf \LsquaredM{}}) condition, which lower bounds the necessary
scaling of a model's history state---the latent variables responsible for
storing past information---for effective long-context modeling. 
We validate the framework and its predictions on transformer and state-space
models. 
Our work provides a principled foundation to understand long-context modeling
and to design more efficient architectures with stronger long-context
capabilities, with potential applications beyond natural language.

\end{abstract}

\section{Introduction}

Large language models (LLMs) have revolutionized natural language processing,
achieving remarkable capabilities across a wide range of tasks
\citep{gpt2,palm,llama,llama2}. Recent advances in large language models,
including ChatGPT \citep{gpt2,gpt4}, Claude, Gemini \citep{gemini,gemini2.5},
Grok, LLaMA \citep{llama2,llama3}, DeepSeek \citep{deepseek, deepseekr1}, and
Qwen \citep{qwen2.5,qwen3} have achieved breakthroughs across diverse tasks,
including code generation, mathematical problem solving, text summarization, and
creative writing
\citep{053b1d7b97eb2c91fc3921d589c160b0923c70b1,fb5c11bbf63884f75d2da615fbf37a3bcfa2bd20,b272513916b45c8517d289d7abee4a53e6832187,9ada8fa11b1cdece31f253acae50b62df8d5f823}.
These models have become increasingly powerful and versatile, pushing the
boundaries of what's possible in natural language processing and marking
significant steps toward artificial general intelligence
\citep{8dbd57469bb32e6d57f23f5e765bf1c9ac8e080c,38179848e2d6a3ad373b1793848816111428ac36,464c3a3512d5bde8078185114f38777843d88256}.

In pushing these advances further, the ability to handle long contexts has
become increasingly crucial. This ability is the key to document-level
understanding, multi-turn dialogue, and complex reasoning. Models like
GPT-o1/o3, Claude Opus, Gemini 2.5 pro, and DeepSeek-R1 often generate extensive
chains of thought, spanning tens of thousands of tokens to solve complex
problems \citep{wei2022chain,wang2024drto1optimizeddeepreasoning}. However,
processing long contexts remains a significant challenge. Despite their success
and expressiveness, transformer architectures suffer from an intrinsic quadratic
computational cost in sequence length, creating challenges for long sequence
generation. Recent advances like DeepSeek have improved per-token efficiency
\citep{deepseek}, yet the fundamental quadratic cost persists.

Although various architectures have been proposed to address the quadratic
scaling
\citep{lineartransformer,ssm,slidingwindow,longformer,longnet,mamba,mamba2,rwkv,ttt,guo2025loglinearattention},
these approaches still struggle with truly long sequences in practice. A
fundamental gap persists in our theoretical understanding of what is necessary
for capturing multi-token long-range dependencies in natural language. Despite
efforts to characterize these dependencies through various statistical measures
\citep{8dd83fa9ee5e7f956800a572c9b975beb3932583,0833b89e988eab7b3c99a99e5da3317da93af0cb,236f10fa4a054e67f7a7cab334f31a80a36781a2,futrell-etal-2019-syntactic}
a theory that can guide practical architecture design remains lacking.

\begin{figure}[t]
    \centering
    \includegraphics[width=\linewidth]{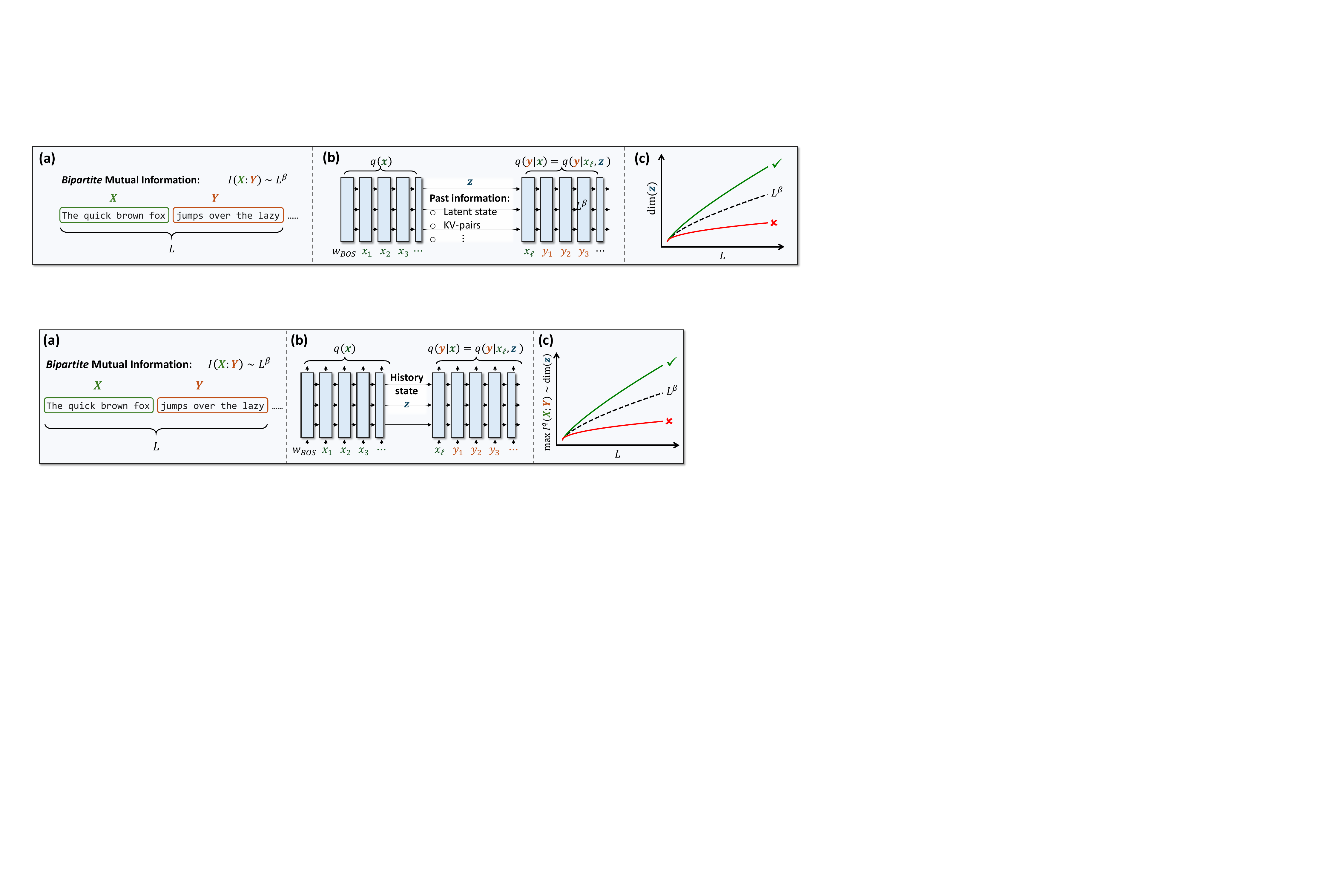}
    \vspace{-15pt}
    \caption{(a) The bipartite mutual information between two text segments
    scales as a power law (sub-volume law) with sequence length $L$.
    (b) In autoregressive models, conditional distributions are parameterized
    through the history state $\bz$, the latent variables that store past
    information.
    Examples of the history state include the recurrent states in state-space
    models or recurrent neural networks, and the key-value pairs in
    transformers.
    (c) The maximum bipartite mutual information a model can express scales with
    the dimensionality of its history state, $\dim(\bz)$. 
    To model long contexts effectively, $\dim(\bz)$ must grow at least as fast
    as the power-law scaling of the true bipartite mutual information.}
    \label{fig:1}
    \vspace{-10pt}
\end{figure}

In this work, we address the challenges of understanding long-context language
modeling through the following contributions (Fig.~\ref{fig:1}).

\begin{enumerate}
    \item We present a universal theoretical framework for autoregressive
    long-context language modeling based on bipartite mutual information.
    \item We demonstrate a bipartite mutual information scaling law in natural
    language and provide reliable empirical validations of power-law scaling
    across diverse natural language datasets using state-of-the-art LLMs.
    \item We derive the \LsquaredM{} condition from this scaling law, lower
    bounding the necessary scaling of a model's history state dimension for
    effective long-context modeling.
    \item We validate our framework and its predictions across transformer and
    state-space model (SSM) architectures on both synthetic and natural language
    datasets of varying lengths.
\end{enumerate}

Our theoretical framework offers crucial insights into understanding an LLM's
capability to model long sequences based on its architectural design. By
identifying the minimum required growth rate of the history state, our work
provides concrete guidance for designing efficient architectures that can
effectively handle long contexts, avoiding the quadratic cost of transformers or
the capacity limitations of fixed-state models, paving the way for future AI
systems.

\section{Related Works} \label{sec:related_works}

{\bf Mutual Information Estimation and Application in Machine Learning}

Mutual information estimation and optimization have been extensively studied in
machine learning, with approaches including variational bounds
\citep{poole2019variationalboundsmutualinformation}, neural estimators
\citep{belghazi2021minemutualinformationneural}, nearest-neighbor methods
\citep{knn,gao2015efficientestimationmutualinformation}, and various upper
bounds \citep{vclub}. It has found wide application in areas such as feature
selection \citep{JMLR:v13:brown12a}, representation learning
\citep{Tschannen2020On}, disentanglement \citep{NIPS2016_7c9d0b1f}, and
generative modeling \citep{hjelm2019learningdeeprepresentationsmutual}.

{\bf Statistical Properties of Natural Language}

Natural language exhibits characteristic statistical scaling behaviors across
different levels of analysis. Zipf's law \citep{Gelbukh2001ZipfAH} describes how
word frequencies decay with their rank, following a power-law distribution.
Heaps' law \citep{Gelbukh2001ZipfAH} characterizes vocabulary growth, showing
that the number of unique words scales sublinearly with text length. Hilberg's
conjecture and its relaxed version posit specific scaling laws for entropy and
bipartite mutual information in natural language, respectively \citep{hilberg}.

{\bf Neural Scaling Laws}

Power-law relationships between model performance, architecture, and
computational requirements have been first empirically observed in neural
networks
\citep{hoffmann2022trainingcomputeoptimallargelanguage,kaplan2020scalinglawsneurallanguage,biderman2023pythiasuiteanalyzinglarge},
with theoretical understanding still being developed
\citep{doi:10.1073/pnas.2311878121,bordelon2024dynamicalmodelneuralscaling},
including recent information-theoretic approach
\citep{nayak2024informationtheorycomputeoptimalsize}. These observations have
guided the development of larger models at fixed context lengths, whereas our
work examines a distinct but complementary question: what determines whether
model architectures can maintain performance as context length increases?

{\bf Universal Prediction and Markov Modeling} 

Recent studies on transformers as
universal predictors
\citep{basu2023transformersuniversalpredictors,zhou2024transformerslearnvariableordermarkov}
show that they can, in principle, model arbitrary variable-order Markov
processes, establishing their theoretical universality in prediction. Our
analysis focuses instead on the information-theoretic scaling that governs how
much past information must be stored to reproduce the mutual-information growth
observed in natural language.

{\bf Architectures for Efficient Long-Context Modeling}

Various approaches have been proposed for processing long sequences.
Architectural innovations targeting quadratic complexity include sparse
attention
\citep{child2019generatinglongsequencessparse,longnet,longformer,NEURIPS2020_c8512d14},
recurrent mechanisms
\citep{dai2019transformerxlattentivelanguagemodels,rae2019compressivetransformerslongrangesequence,sukhbaatar2019adaptiveattentionspantransformers},
and alternative formulations
\citep{lineartransformer,he2024hmthierarchicalmemorytransformer,ssm,mamba,mamba2,rwkv,ttt,gu2022efficientlymodelinglongsequences,beck2024xlstmextendedlongshortterm,de2024griffinmixinggatedlinear,guo2025loglinearattention}.
Efficient attention implementations like Flash Attention
\citep{dao2022flashattentionfastmemoryefficientexact,dao2023flashattention2fasterattentionbetter,shah2024flashattention3fastaccurateattention},
Lightning Attention \citep{qin2024lightningattention2freelunch}, and Paged
Attention \citep{kwon2023efficientmemorymanagementlarge} have improved per-token
computational efficiency despite maintaining the underlying complexity scaling.

{\bf Long-Form Reasoning and Context Utilization}

Chain-of-thought prompting \citep{wei2022chain} and scratchpad methods
\citep{nye2021workscratchpadsintermediatecomputation} demonstrate the importance
of extended context for complex reasoning tasks, emphasizing the urgent need for
effective long-range dependency modeling.

{\bf Information Theory and Physics-Inspired Approaches}

Recent work has demonstrated how information-theoretic principles and
physics-inspired approaches can guide machine learning
\citep{7133169,9082644,10835527}, leading to novel architectures
\citep{PhysRevLett.122.226401,PhysRevResearch.5.013216,wang2021spacetimeneuralnetworkhigh,chen2022simulating21dlatticequantum,PhysRevB.106.L081111,Wu_2023,NEURIPS2023_01772a8b,dugan2024occamllmfastexactlanguage},
training methods
\citep{Stokes2020quantumnatural,chen2024teng,chen2024quantaefficienthighrankfinetuning},
and broad applications
\citep{RevModPhys.91.045002,Lee2020,PhysRevLett.128.090501,moro2024multimodallearningmaterials,chen2024teng,Choi2024}.

\section{Preliminaries} \label{sec:ann} {\bf Mutual Information}

Mutual information $I(X;Y)$ quantifies the statistical dependence between random
variables $X$ and $Y$, defined as $I(X;Y) = D_{KL}(p_{XY} || p_X \otimes p_Y)$,
where $D_{KL}(\cdot || \cdot)$ is the Kullback--Leibler (KL) divergence, and
$p_{XY}$ is the joint distribution of $X$ and $Y$. For discrete random
variables, mutual information permits equivalent formulations as
\begin{equation}
\begin{aligned}
I(X;Y) = H(X) + H(Y) - H(XY) 
= H(X) - H(X|Y) 
= H(Y) - H(Y|X),
\end{aligned}
\end{equation}

where $H(\cdot)$ is the (Shannon) entropy and $H(\cdot|\cdot)$ is the
conditional entropy. 
This definition naturally extends to collections of random variables:
$I(X_{1:m};Y_{1:n})$, with $X_{i:j}$ denoting the sequence $(X_i,\ldots,X_j)$.
For notational convenience, we will use boldface notation $\bX:=X_{i:j}$ when
the indices are clear from context. Similarly, we will drop the index of a
single variable $X:=X_i$ when convenient.

{\bf Autoregressive Neural Networks}

Modern LLMs predominantly employ autoregressive neural architectures. An
autoregressive neural network models a sequence of conditional probability
distributions over tokens $\{q(w_i|w_{1:i-1},w_{\text{BOS}})\}_{i=1}^L$, where
$w_{\text{BOS}}$ is the beginning-of-sequence token.
Throughout this paper, we use $q$ to denote model-generated probability
distributions (and sometimes the model itself) and $p$ to denote the true
underlying distributions. Upper case letters denote random variables, and lower
case letters denote specific values or realizations of these random variables.
These conditional distributions jointly model the probability for a sequence of
tokens given a prefix as 
\begin{equation}
q(w_{\ell:L}|w_{1:\ell-1}, w_{\text{BOS}}) = \prod_{i=\ell}^L q(w_i|w_{1:i-1},w_{\text{BOS}}).
\end{equation}
When $\ell=1$, this reduces to the distribution of unconditional generation
$q(w_{1:L}|w_{\text{BOS}})$. During inference, tokens are sampled sequentially
from these conditional distributions to generate text or respond to prompts.

For a complete list of notation and conventions used throughout this paper, we
refer the reader to Appx.~\ref{app:notation}.

\begin{figure}[ht]
    \centering
    \includegraphics[width=\linewidth]{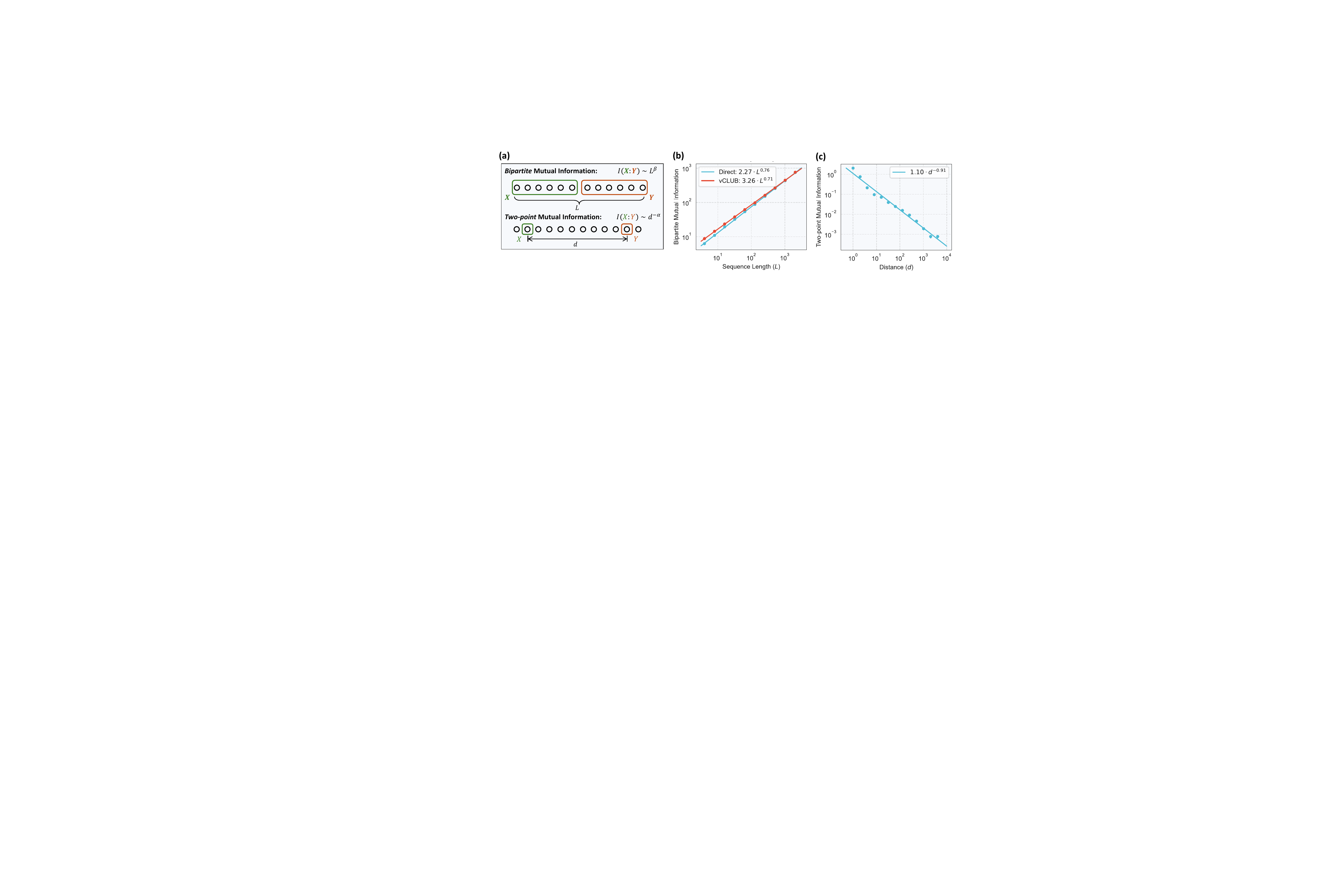}
    \vspace{-15pt}
    \caption{{(a)} Illustration of bipartite and two-point mutual information.
    The \emph{bipartite} mutual information measures statistical dependence
    between two adjacent segments within a text block of length $L$, whereas the
    two-point mutual information measures the dependence between two tokens
    separated by a distance $d$. {(b)} Estimates of bipartite mutual information
    using LLaMA 3.1 405B model \citep{llama405} on \PG{} dataset \citep{pg19} of
    pre-1919 books. {(c)} Estimates of two-point mutual information on \PG{}
    dataset. See Appx.~\ref{app:mi_other_llm},~\ref{app:mi_other_l},
    and~\ref{app:two-point} for additional results.}
    \vspace{-5pt}
    \label{fig:real_data_measurement}
\end{figure}

\section{Mutual Information Scaling Laws}

\subsection{Bipartite Mutual Information as Predictive Information}

While classical scaling laws in natural language, such as Zipf's and Heaps'
laws, primarily address token-level statistics, a deeper understanding of
language modeling necessitates analyzing dependencies between entire text
segments. A central challenge in modeling language effectively is to
characterize how information is carried over from an existing block of text,
$\bX$, to inform the generation of a subsequent block, $\bY$. The
\emph{bipartite} mutual information between such adjacent blocks directly
quantifies this inter-segment information transfer, emerging as a particularly
revealing measure.

\begin{definition}[\emph{Bipartite} Mutual Information
    {[Fig.~\ref{fig:real_data_measurement}(a)]}] For a consecutive sequence of
    tokens (random variables) $W_{1:L}$ of length $L$, consider a bipartition of
    the tokens: $X_{1:\ell} := W_{1:\ell}$ and $Y_{1:L-\ell} := W_{\ell+1:L}$.
    The bipartite mutual information is the mutual information between the two
    parts $I_{\ell;L}^{\mathrm{BP}} := I(X_{1:\ell};Y_{1:L-\ell})$.
\end{definition}

The role of bipartite mutual information in quantifying this predictive
relationship is formally illuminated by decomposing the entropy of the
subsequent block $\bY$:
\begin{equation}
    H(\bY) = H(\bY|\bX) + I(\bX;\bY) = H(\bY|\bX) + I^{\mathrm{BP}}.
\end{equation}
This decomposition shows that the total information in $\bY$ (its entropy
$H(\bY)$) consists of two distinct components: new information unique to $\bY$
given $\bX$ (the conditional entropy $H(\bY|\bX)$), and information that $\bY$
shares with $\bX$ (the bipartite mutual information $I^{\mathrm{BP}}$).
Consequently, $I^{\mathrm{BP}}$ precisely measures the amount of information
from the preceding block $\bX$ that is predictive of the next block $\bY$, and
therefore, bipartite mutual information is also referred to as predictive
information \citep{bialek1999predictiveinformation}.

Despite its crucial role in quantifying predictive information, this form of
mutual information in language has remained relatively underexplored. This
research gap is primarily due to two factors: the absence of a comprehensive
theory of natural language that would permit a direct calculation, and the
substantial challenges in empirically measuring entropy and mutual information
for high-dimensional distributions from samples. 

Existing literature offers differing perspectives on its scaling properties. On
one hand, analogies drawing from critical physical systems
\citep{Ebeling_1994,EBELING1995233,doi:10.1142/S0218348X02001257,shen2019mutualinformationscalingexpressive,e19070299,powerLawsUniversality}---often
based on two-point mutual information scaling (discussed later)---suggest that
bipartite mutual information should scale logarithmically with sequence length.
On the other hand, research in computational linguistics has proposed that it
follows power-law growth \cite{relaxedhilbergconjecture}, a behavior often
referred to as the sub-volume law (these terms are used interchangeably in this
paper). Previous empirical efforts to measure such scaling have been constrained
by methodological biases and the curse of dimensionality
\citep{hilberg,relaxedhilbergconjecture,mine_mi_measure}. Although existing
evidence tends to favor sub-volume law growth, these limitations have prevented
a definitive characterization. In Sec.~\ref{sec:measure_mi}, we address these
challenges by leveraging state-of-the-art LLMs as density estimators,
establishing clear power-law scaling for bipartite mutual information across
diverse datasets.

\subsection{Two-point Mutual Information}

Before presenting our main results concerning bipartite mutual information
scaling, it is instructive to discuss two-point mutual information. This measure
has conventionally been used to assess long-range dependencies in natural
language, and its scaling properties are relatively well understood.

\begin{definition}[\emph{Two-point} Mutual Information
{[Fig.~\ref{fig:real_data_measurement}(a)]}] The two-point mutual information
measures the mutual information between two tokens (random variables) $X$ and
$Y$ separated by a distance $d$: $I_d^{\mathrm{TP}} = I(X;Y)$.
\end{definition}
Specifically, two-point mutual information has been observed to follow a
power-law decay, $I_d^{\mathrm{TP}}\sim d^{-\alpha}$
\citep{Ebeling_1994,EBELING1995233,doi:10.1142/S0218348X02001257,shen2019mutualinformationscalingexpressive}.
This characteristic decay has prompted arguments that natural language shares
structural properties with critical physical systems, which exhibit similar
two-point correlation behavior \citep{e19070299,powerLawsUniversality}. However,
we contend that such analogies, while offering certain insights, can be
misleading when assessing the full complexity of multi-token dependencies
crucial for language modeling. The limitations of two-point mutual information
in this regard, and why it provides an incomplete characterization for this
task, will be detailed in Sec.~\ref{sec:distiction} and
Appx.~\ref{app:two-point-failure}. Our present discussion of two-point mutual
information serves primarily to contrast it with the bipartite measure that is
central to our work.

\subsection{Empirical Verification of Mutual Information Scaling Laws}
\label{sec:measure_mi}

{\bf Bipartite Mutual Information.} Measuring bipartite mutual information
presents significant challenges without access to the underlying probability
distribution $p$. Traditional estimation methods face severe limitations in our
setting: K-nearest neighbor estimators
\citep{gao2015efficientestimationmutualinformation} and neural estimators like
MINE \citep{belghazi2021minemutualinformationneural} and InfoNCE
\citep{oord2019representationlearningcontrastivepredictive} struggle with the
high dimensionality of long text sequences, with errors that increase rapidly as
sequence length grows. Additionally, neural estimators require substantial
training on large amounts of data to learn representations of natural language
distributions, especially for long sequences.  
Fortunately, recent advances in LLMs allow us to circumvent training our own
density estimators by offering high-quality approximations $q$ to these
distributions (see Appx.~\ref{app:mi_estimation_methods} for additional
discussions). As autoregressive models, LLMs enable efficient computation of
conditional probabilities (Sec.~\ref{sec:ann}) and their associated
cross-entropies (negative log-likelihoods):
\begin{equation}
H(p_{\bY|\bX}, q_{\bY|\bX}):= -\expect_{p_{\bX\bY}} \log q(\bY|\bX),
\end{equation}
where the expectation is taken over samples from the true underlying
distribution $p_{\bX\bY}$. The cross-entropy provides an upper bound to the true
entropy:
\begin{equation} \label{eq:nll_kl_entropy}
H(p_{\bY|\bX}, q_{\bY|\bX})=D_{KL}(p_{\bY|\bX}||q_{\bY|\bX})+H^p(\bY|\bX) \ge H^p(\bY|\bX),
\end{equation}
where the conditional cross-entropy and KL divergence implicitly average over
$p_{\bX}$, and $H^p$ (or $H^q$) denotes the entropy computed with respect to
distribution $p$ (or $q$).

Using these properties, we can construct a direct estimator for bipartite mutual
information:
\begin{equation} \label{eq:mi_direct}
\begin{aligned}
    I_{\ell;L}^{\mathrm{BP},\mathrm{direct}} =\expect_{p_{\bX\bY}} \left[ \log q(\bY|\bX) - \log q(\bY) \right]
    = I^{p}(\bX;\bY) + \varepsilon(p, q),
\end{aligned}
\end{equation}
where $I^{p}(\bX;\bY)$ denotes mutual information with respect to $p$, and
$\varepsilon(p, q) = D_{KL}(p_{\bY}||q_{\bY}) -
D_{KL}(p_{\bY|\bX}||q_{\bY|\bX})$. While this estimator no longer provides a
bound, it preserves the key property that $\varepsilon(p, q)\rightarrow 0$ as
$q\rightarrow p$.

We note that this estimation method faces a specific challenge with modern LLMs:
they model $q(w_i|w_{1:i-1}, w_{BOS})$ rather than $q(w_i|w_{1:i-1})$, where
$w_{BOS}$ denotes the BOS token. When sampling from the dataset, we can ensure
$\bX$ starts at sentence beginnings, making $q(\bY|\bX,w_{BOS})\equiv
q(\bY|\bX)$. However, $\bY$ may start mid-sentence, creating a mismatch where
$q(\bY) \neq q(\bY|w_{BOS})$. This introduces errors in estimating $H(p_{\bY},
q_{\bY})$. We address this using $n$-gram corrections for the first two tokens,
which are the primary source of this bias (see Appx.~\ref{app:mi_direct}).

To circumvent issues with estimating $q(\bY)$, we also employ the vCLUB
estimator \cite{vclub}:
\begin{equation} \label{eq:mi_vclub}
    I_{\ell;L}^{\mathrm{BP},\mathrm{vCLUB}} = \expect_{p_{\bX\bY}} \log q(\bY|\bX) - \expect_{p_{\bX} \otimes p_{\bY}} \log q(\bY|\bX),
\end{equation}
where the second term can be calculated by shuffling the second halves of
samples in the dataset. Analysis in \cite{vclub} shows that vCLUB provides an
upper bound on the true bipartite mutual information when $q$ closely
approximates $p$. Even when $q$ deviates moderately from $p$, though the upper
bound property may not hold, vCLUB continues to provide reliable estimates of
the true bipartite mutual information.

Our empirical analysis in Fig.~\ref{fig:real_data_measurement}(b) focuses on
equal-length partitions of $\bX$ and $\bY$ ($\ell=L/2$), where the bipartite
mutual information tends to maximize for fixed $L$'s. Nevertheless, the same
analysis can be carried out using other partitions where similar results can be
obtained (with the results in Appx.~\ref{app:mi_other_l}). Using both the
bias-corrected direct estimator [Eq.\eqref{eq:mi_direct}] and vCLUB estimator
[Eq.~\eqref{eq:mi_vclub}], we measure scaling on the \PG{} dataset\footnote{We
avoid the \Books{} dataset due to copyright infringement concerns.} \citep{pg19}
(a collection of books before 1919), employing the LLaMA 3.1 405B model
\citep{llama405} as density estimator $q$. All measurements robustly demonstrate
a clear power-law scaling that extends across thousands of tokens. Additional
measurements on \Wikipedia{} \citep{wikidump} and using additional LLMs, along
with varying $\ell/L$ ratios, can be found in Appx.~\ref{app:mi_other_llm}
and~\ref{app:mi_other_l}. We note that both estimators likely underestimate the
true exponent $\beta$ (see Appx.~\ref{app:underestimate} for discussions).

{\bf Two-point Mutual Information.} For completeness, we also measure two-point
mutual information scaling on the same datasets, confirming the expected
power-law decay [Fig.~\ref{fig:real_data_measurement} (c)]. Detailed
methodologies for these measurements are provided in Appx.~\ref{app:two-point}
and~\ref{app:two-point-bias-correction}.

\subsection{Failures of Two-point Mutual Information} \label{sec:distiction}

As previously noted, while two-point mutual information is easier to measure and
more frequently studied in existing literature, it often fails to adequately
capture the long-range multi-token dependencies crucial for natural language
modeling. When modeling language, our primary concern is the accurate prediction
of future tokens given a preceding context, i.e., $q(w_{\ell:L}|w_{1:\ell-1},
w_{BOS})$. Effective modeling of this conditional distribution necessitates a
clear understanding of the multi-token dependencies between the context
$w_{1:\ell-1}$ and the subsequent tokens $w_{\ell:L}$. It is important to
recognize that this multi-token dependency cannot always be accurately
represented by a simple aggregation of pairwise (two-point) interactions; such
an approach can be insufficient or even misleading in certain contexts. The
following examples illustrate these potential limitations, and we provide more
formal derivations in Appx.~\ref{app:two-point-failure}.

Consider a simple distribution where all tokens must be identical:
$p(x_1,x_2,\dots,x_L) = \mathbbm{1}(x_1=x_2=\cdots=x_L) / M$, where
$\mathbbm{1}(\cdot)$ is the indicator function that evaluates to 1 when the
condition is satisfied and 0 otherwise, and $M$ is the vocabulary size. This
distribution permits a Markov chain construction, as $p(x_1) = 1/M$ and
$p(x_i|x_{1:i-1}) = \mathbbm{1}(x_i=x_{i-1})$, thus possessing a simple
token-to-token dependency structure. Despite this inherent simplicity, the
two-point mutual information suggests a misleadingly strong ``long-range''
dependency: it maintains a large, constant value of $I_{d}^{\mathrm{TP}}=\log M$
regardless of the distance $d$, significantly larger than the decaying two-point
mutual information typically observed in natural languages. In contrast,
bipartite mutual information correctly reflects this simple dependency
structure, with $I_{\ell;L}^{\mathrm{BP}}=\log M$ remaining constant for any
choice of $\ell$ and $L$. This indicates that any two segments share exactly the
same amount of information ($\log M$), which is no more than the information
shared between just two adjacent tokens, accurately capturing the limited nature
of the dependency.

For a more realistic setting, we refer to Appx.~\ref{app:gaussian} for a
discussion of two families of multivariate Gaussian distributions of varying
lengths (details of their construction are in
Appx~\ref{app:gaussian_construction}). Notably, both families exhibit identical
power-law decay in their two-point mutual information when measured between
variables at maximum separation. However, their bipartite mutual information
scaling differs dramatically: one scales as $L^{\beta}$, akin to natural
language, while the other scales as $\log L$, similar to that observed in
critical physical systems. This disparity further underscores that two-point
mutual information alone may be insufficient to distinguish between systems with
fundamentally different long-range correlational structures.

\section{Long-Context Language Modeling (\LsquaredM) Condition}
\label{sec:theory}

Having established bipartite mutual information as a crucial tool for measuring
long-range dependencies, we analyze how a model's capacity to handle long
contexts fundamentally depends on its ability to store past information, using
bipartite mutual information scaling as our theoretical framework. Intuitively,
to model natural language effectively, a model must be able to capture all
dependencies between past and future tokens. Since these dependencies (measured
by bipartite mutual information) grow with sequence length, the model's state
capacity for storing past information (the history state) must necessarily grow
as well. We formalize this intuition through the \LsquaredM{} condition and
explore its implications in detail throughout this section.

\subsection{Theoretical Derivations}

To analyze how models handle long-range dependencies, we first formalize the
notion of \emph{history state}.

\begin{definition} \label{def:model_state_z} Consider a sequence of tokens
    $w_{1:L}$. Denote $x_{1:\ell}:=w_{1:\ell}$ and $y_{1:L-\ell}:=w_{\ell+1:L}$.
    Autoregressive neural networks parameterize conditional probabilities by
    first encoding the input tokens $x_{1:\ell-1}$ into a set of latent
    intermediate variables $\bz_\ell= \boldsymbol{f}(x_{1:\ell-1})$ before
    outputting the conditional probabilities as $q(y_{1:L-\ell}|x_{1:\ell}):=
    q(y_{1:L-\ell}|x_\ell,\bz_\ell)$.\footnote{We separate $x_\ell$ from
    $\bz_\ell$ to accurately reflect its distinct role as the current input
    token in autoregressive models, though including it in $\bz_\ell$ would not
    affect the main results of this paper.} We define the \emph{history state}
    as the smallest set of such latent intermediate variables that fully
    characterizes the model's output conditional probability.
    [Fig.~\ref{fig:1}(b)].
\end{definition}

As illuminating examples, the history state corresponds to the recurrent state
in RNNs and SSMs after processing token $w_{\ell-1}$, and to the key-value pairs
up to token $w_{\ell-1}$ for transformers (see Appx.~\ref{app:history_state}).
Generally, the history state $\bz_\ell$ is the smallest hidden state responsible
for caching all historical information.

The following theorem shows that this history state upper bounds a model's
capacity to capture bipartite mutual information:
\begin{theorem} \label{thm:mi_to_z} The bipartite mutual information that a
model can capture is bounded by the size of its history state:
\begin{equation}
I_{L/2;L}^{\mathrm{BP},q} \leq C \cdot \dim(\boldsymbol{z}_{L/2}) + \log(M)
\end{equation}
where $C$ is some constant and $M$ denotes the vocabulary size.
\end{theorem}

\begin{proof}
    This theorem admits multiple independent proofs under different mild and
    practical assumptions. See Appx.~\ref{app:proof} for details.
\end{proof}

We now use this bound to analyze when architectures can maintain performance as
sequence length increases. Consider a series of natural language datasets
$\{W_{1:L}\}_{L=1}^\infty$ of different lengths, which can be thought of as
truncations of an ideal infinite-length dataset.

\begin{definition}
A model $q$ is \emph{MI-capable} if the maximum bipartite mutual information it
can express satisfies $\max\limits_\btheta I_{L/2;L}^{\mathrm{BP},q_\btheta}
\geq I_{L/2;L}^{\mathrm{BP}}$ for any sequence length $L$, where the maximum is
taken over all model parameters $\btheta$.
\end{definition}

Since a model's ability to capture mutual information is bounded by its history
state dimension, we immediately obtain\footnote{See Appx.~\ref{app:notation} for
conventions on asymptotic notations.}:

\begin{framed}
\begin{theorem}[\LsquaredM{} Condition for Single Models]
For a model to be \emph{MI-capable} across all sequence lengths, its history
states $\bz_{L/2}^{q}$ must satisfy $\dim(\bz_{L/2}^{q}) \succsim
I_{L/2;L}^{\mathrm{BP}} \sim L^\beta$.
\end{theorem}
\end{framed}

\begin{proof}
We prove by contrapositive. By Thm.~\ref{thm:mi_to_z}, if $\dim(\bz_{L/2}) \prec
I_{L/2;L}^{\mathrm{BP}}$, then $\max\limits_\btheta I_{L/2;L}^{\mathrm{BP},
q_\btheta}\prec I_{L/2;L}^{\mathrm{BP}}$, implying there exists some $L$ where
$\max\limits_\btheta I_{L/2;L}^{\mathrm{BP}, q_\btheta} <
I_{L/2;L}^{\mathrm{BP}}$, violating MI-capability.
\end{proof}

For some architectures, a single fixed-size model may not satisfy this condition
across all sequence lengths. In such cases, we can extend our framework to
families of models where model size grows with sequence length. Consider a
series of models $\{q_L\}_{L=1}^\infty$ of the same architecture, where model
size may increase with $L$.

\begin{definition}
A series of models $\{q_L\}_{L=1}^\infty$ is \emph{MI-capable} if the maximum
bipartite mutual information each model can express satisfies
$\max\limits_{\btheta_L} I_{L/2;L}^{\mathrm{BP},q_{L,\btheta_L}} \geq
I_{L/2;L}^{\mathrm{BP}}$ for its corresponding sequence length $L$, where the
maximum is taken over all parameters $\btheta_L$ of model $q_L$.
\end{definition}

\begin{framed}
\begin{theorem}[\LsquaredM{} Condition for Model Series]
For a series of models $\{q_L\}_{L=1}^\infty$ to be \emph{MI-capable}, the
history states $\bz_{L/2}^{q_L}$ of each model must satisfy:
$\dim(\bz_{L/2}^{q_L}) \succsim I_{L/2;L}^{\mathrm{BP}} \sim L^\beta$.
\end{theorem}
\end{framed}

Note that an MI-capable single model trivially induces an MI-capable series when
applied to all sequence lengths, though the converse is not true.

\subsection{Implications to Common LLM Architectures}

We can now apply our framework to analyze whether different architectures
satisfy the \LsquaredM{} condition and thus can capture long-range dependencies
as sequence length grows.

In transformer-based models (excluding sparse attention and linear attention
variants), the history state consists of stored key-value pairs for all previous
tokens. Even with fixed model size, these key-value pairs grow linearly with
sequence length: $\dim(\bz_{L/2}^{q}) \sim L \succsim L^\beta$. This means a
single transformer model naturally satisfies the \LsquaredM{} (single model)
condition across all sequence lengths, notwithstanding the quadratic
computational cost.

In contrast, SSMs, RNNs, and linear attention models, despite being celebrated
for their ``infinite'' context length and linear complexity, cannot satisfy the
\LsquaredM{} condition with a single fixed-size model. Their history state
dimension remains constant regardless of sequence length, and our theory
demonstrates that this constant-size state cannot capture the growing mutual
information. However, these architectures can achieve MI-capability
(model-series) through a series of models $\{q_L\}_{L=1}^\infty$ where model
size, and thus history state dimension, increases with sequence length. This
requirement effectively offsets their computational efficiency advantage when
modeling long sequences.\footnote{And a new model must be trained for each
sequence length, which can be prohibitively expensive.}

For other architectures, such as sparse attention models and log-linear models,
we can similarly analyze their history state scaling to determine whether they
satisfy the \LsquaredM{} condition as single models or require a series of
growing models. Crucially, any architecture must exhibit power-law growth in its
history state dimension with sequence length in order to truly satisfy the
single-model \LsquaredM{} condition.

We note that the \LsquaredM{} condition addresses a model's capacity to capture
long-range dependencies, not its overall language modeling capability. It is a
necessary but not sufficient condition: architectures that fail to satisfy it
will have inherent limitations at longer sequences, while satisfying it does not
guarantee effective language modeling. As discussed in
Sec.~\ref{sec:related_works}, the \LsquaredM{} condition is also distinct from
neural scaling laws, which typically study how model performance scales with
model size, dataset size, and compute budget at a \emph{fixed} sequence length.

\section{Empirical Verification}

\begin{figure}[t]
    \centering
    \includegraphics[width=\linewidth]{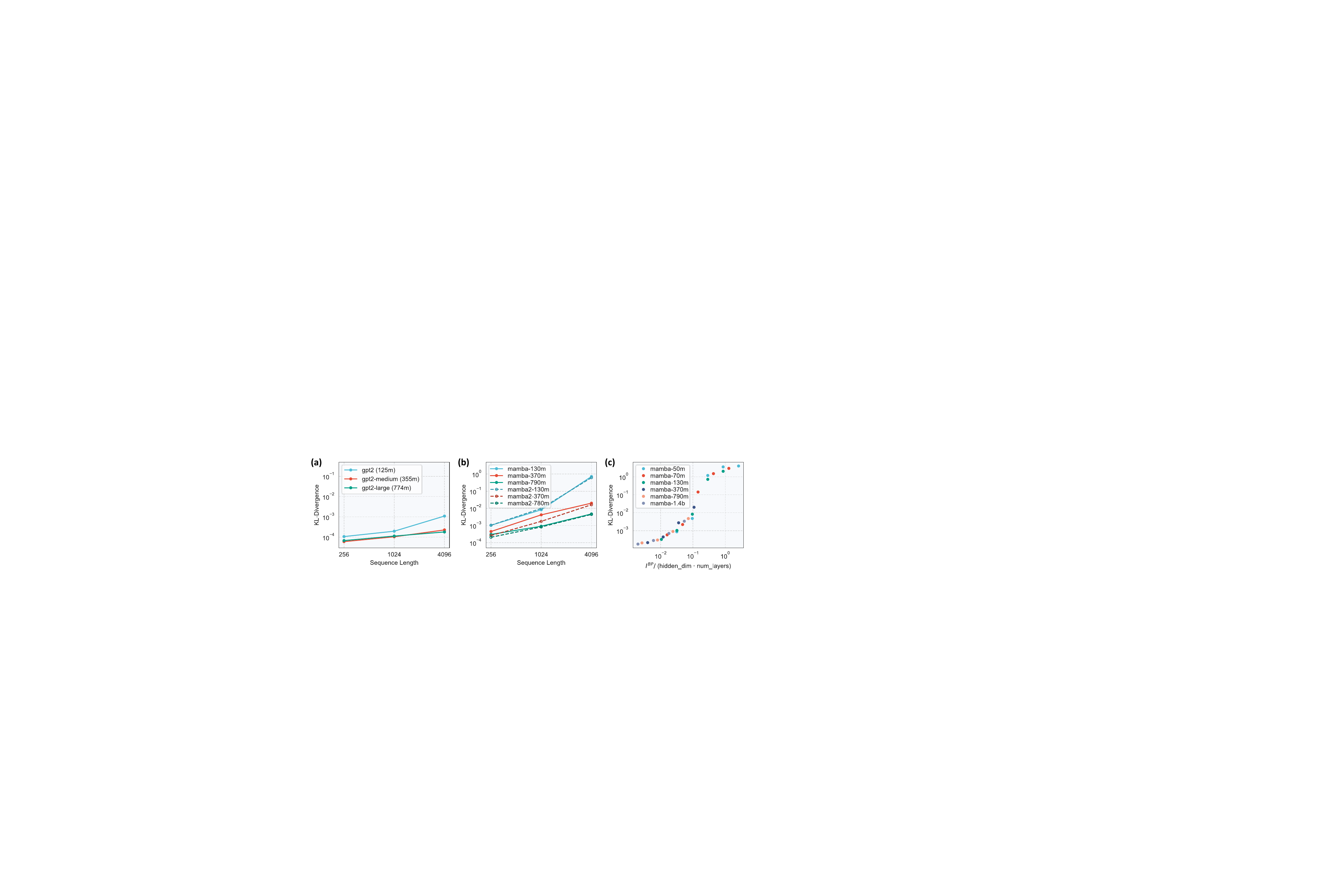}
    \vspace{-15pt}
    \caption{Evaluation of KL-divergence across model architectures trained on synthetic data that satisifes the bipartite mutual information scaling. (a, b) Average KL-divergence per token for models trained on different sequence lengths. (c) Average KL-divergence per token as a function of the ratio between bipartite mutual information and Mamba recurrent state sizes.}
    \vspace{-10pt}
    \label{fig:syn_kl_div}
\end{figure}

{\bf Sub-volume-law Gaussian Test.} We first validate our theory using a
synthetic dataset comprising a family of multivariate Gaussian distributions
(see Appx.~\ref{app:gaussian} for details). This distribution family closely
mimics the scaling of both bipartite and two-point mutual information observed
in natural language, while crucially allowing for the efficient calculation of
conditional probabilities and KL divergences---calculations that would be
intractable with real-world natural language datasets. Furthermore, the
synthetic data enables an isolated assessment of a model's ability to handle
long sequences, without interference from its capacity to understand the
semantic meanings of natural language.

In Fig.~\ref{fig:syn_kl_div}, we present the average per-token KL divergence
(defined in Appx.~\ref{app:experiment_eval_metric}) for GPT2, Mamba, and Mamba2
models, serving as representative transformer and SSM architectures. Panels (a)
and (b) show that GPT2 maintains consistent KL divergence across different
sequence lengths. In contrast, smaller Mamba and Mamba2 models exhibit
increasing difficulty with longer contexts, necessitating substantially larger
model sizes to achieve comparable performance at a sequence length of 4096.
Panel (c) offers direct confirmation of our theoretical framework: it plots KL
divergence against the ratio of bipartite mutual information to the recurrent
state size for Mamba models of varying configurations. For this, we varied
sequence lengths from 64 to 16,384 and model sizes from 50M to 1.4B parameters.
The KL divergence values from these diverse configurations remarkably collapse
onto a single curve, demonstrating that model performance depends only on the
ratio $I^{\mathrm{BP}}/\dim(\bz)$. This finding precisely confirms our theory
that for effective long-context modeling, a model's history state size must
scale at least as fast as the bipartite mutual information present in the data. 

These findings have important implications for modeling very long sequences.
Extrapolating from the measured scaling in Fig.~\ref{fig:real_data_measurement}
(which likely underestimates the true exponent), the bipartite mutual
information for a sequence of one million tokens could exceed 60,000 nats. Our
results in Fig.~\ref{fig:syn_kl_div}(c) suggest that maintaining low KL
divergence at such bipartite mutual information levels would require recurrent
state dimensions approaching one million.

\begin{figure}[t]
    \centering
    \includegraphics[width=\linewidth]{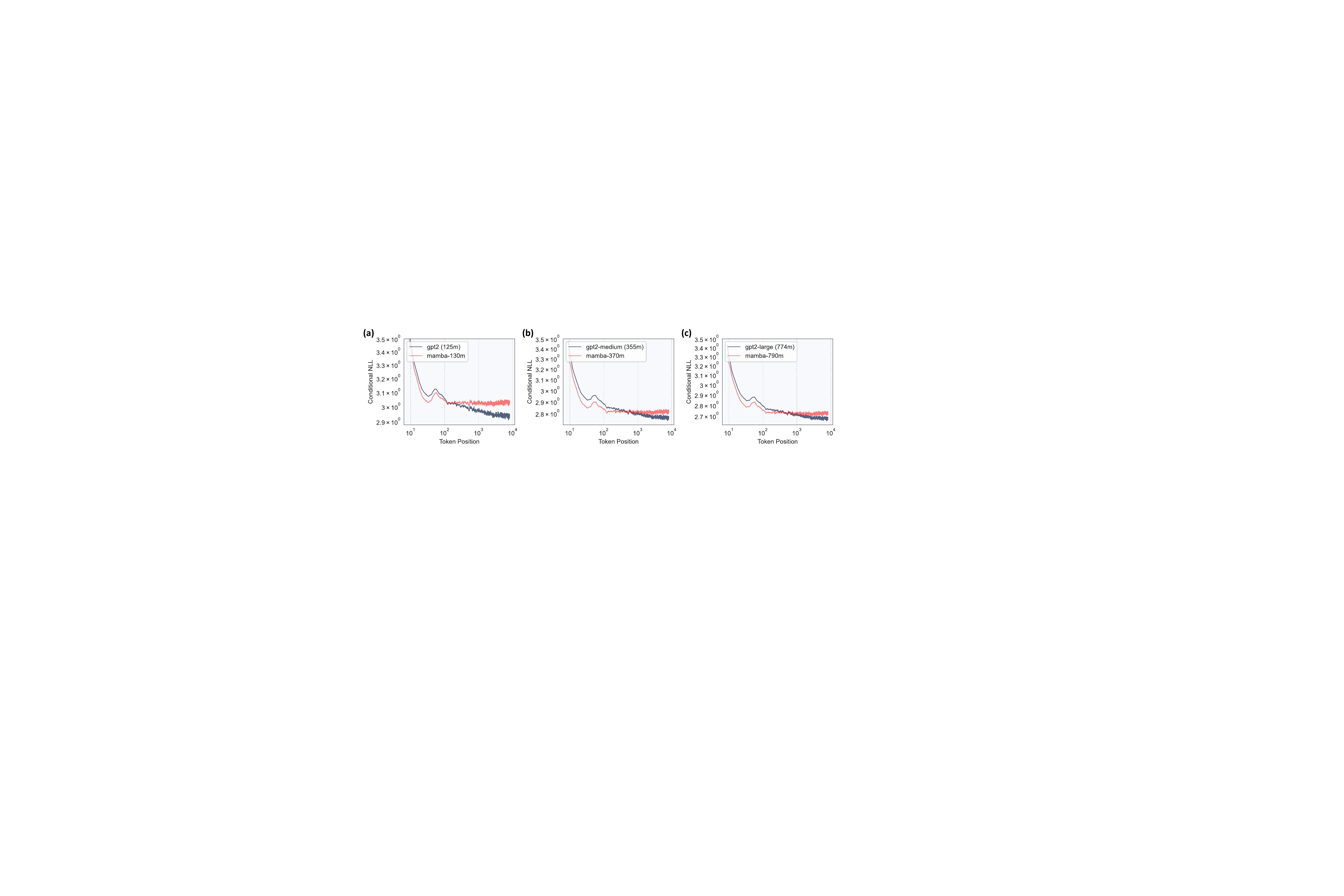}
    \vspace{-15pt}
    \caption{Position-wise conditional negative log likelihood (NLL) evaluation
    for models trained on 8192-token sequences on the \PG{} dataset
    \citep{pg19}.}
    \vspace{-10pt}
    \label{fig:real_nll}
\end{figure}

{\bf \PG{} Test.} We then extend our analysis to the \PG{} dataset \citep{pg19},
a high-quality collection of pre-1919 books exhibiting long contextual
dependencies.

In Fig.~\ref{fig:real_nll}, we show the position-wise conditional negative log
likelihood (NLL) of models trained on the \PG{} dataset \citep{pg19} with
8192-token sequences, where calculating KL-divergence is not feasible. Note
that, unlike conditional KL divergence, conditional NLL naturally decreases with
token position (see Appx.~\ref{app:experiment} for details). Two key patterns
emerge from this experiment: First, Mamba models typically outperform GPT2
models of comparable size at early token positions, but this advantage
diminishes and eventually reverses at later positions. Most notably, Mamba's NLL
tends to plateau beyond certain positions unless the model size is increased,
while GPT2's NLL continues to improve. Second, the performance gap between Mamba
and GPT2 narrows with increasing model size. Both observations align with our
theoretical predictions: since Mamba's history state size remains fixed
regardless of sequence position, its performance inevitably degrades beyond a
certain token position unless model size increases. As model size grows, the
history state size also increases, eventually becoming sufficient to capture the
mutual information present in 8192-token sequences.

We note that Mamba's linear computational complexity can make larger Mamba
models practically more efficient than smaller transformers. Our results should
not be interpreted as suggesting Mamba's architectural inferiority. Rather, they
demonstrate how different architectures handle long sequences differently, and
that a model's capacity for capturing long-range dependencies aligns with our
theoretical \LsquaredM{} framework, regardless of the architecture.

Additional experimental results can be found in Appx.~\ref{app:add_experiment}.

\section{Discussion}

The \LsquaredM{} condition establishes a fundamental relationship between the
information structure of data and architectural requirements. This relationship
manifests differently across architectures: transformers with linearly growing
key-value caches naturally satisfy the condition as single models (given our
measured sublinear mutual information scaling with $\beta < 1$), though at
quadratic computational cost, while SSMs and similar fixed-state architectures
require model size to scale with sequence length to achieve comparable mutual
information capability. 

Interestingly, transformers appear to \emph{over-provision} their history state
relative to the measured mutual information scaling: their linear growth exceeds
the sublinear ($L^\beta$ with $\beta < 1$) scaling we observe. This observation
provides a clear goal for future architecture design. Although it remains
unclear whether the over-provisioning is necessary for other aspects of language
modeling beyond pure information storage, the gap between the linear growth of
transformers and the $L^\beta$ requirement suggests a concrete target:
architectures that precisely match the required sublinear scaling could
potentially achieve substantially improved efficiency while maintaining the
capacity to capture long-range dependencies.

Our framework applies to autoregressive language models, which encompass the
vast majority of widely-used LLMs. While diffusion-based language models
represent an alternative generative paradigm, they typically still operate
autoregressively at a higher level of granularity, making our framework
applicable in practice. Extending our framework to hybrid architectures that
combine different mechanisms represents an important research direction that
could unify our understanding of how diverse architectural choices affect
long-context capabilities. Applying our framework to other sequential domains,
such as biological sequences like proteins or DNA, or computer code, also
presents a particularly promising direction, as different mutual information
scaling behaviors in these domains could provide a principled explanation for
the observed differences in model requirements across domains.

\section{Conclusion}

We establish a bipartite mutual information scaling law that characterizes
long-range dependencies in natural language and introduce the \LsquaredM{}
condition, which lower bounds the necessary scaling of a model's history state
for effective long-context modeling. By identifying the minimum required growth
rate of the history state, our work provides a principled foundation for
understanding how different architectures handle long contexts. This framework
establishes concrete, information-theoretical, and data-driven targets that
could guide the design of architectures balancing computational efficiency with
the capacity to capture long-range dependencies in natural language and
potentially beyond.

\newpage

\section*{Limitations}

Our theoretical framework specifically examines models' capacity to capture
long-range dependencies through the lens of bipartite mutual information and
does not address other aspects of language modeling, such as reasoning
capabilities or world knowledge. The \LsquaredM{} condition establishes
necessary but not sufficient conditions for effective long-context modeling.
Understanding how this theoretical capacity translates to actual downstream task
performance remains an important open question. The relationship likely depends
on additional factors including optimization dynamics, architectural inductive
biases, and task-specific requirements. Systematic evaluation across diverse
long-context benchmarks represents a crucial next step to clarify these
relationships and identify any gaps between theoretical capability and practical
performance.

While our empirical validation on synthetic Gaussian distributions with
controlled mutual information scaling provides clean verification of the
theoretical predictions, it may not capture all complexities present in natural
language. Our theory focuses on autoregressive language models, which remains
broadly applicable as even diffusion-based approaches typically employ
autoregressive generation for extended sequences in practice. Nonetheless,
exploring whether similar information-theoretic principles govern fundamentally
different generative paradigms or multimodal models represents an interesting
direction for future work.

The methodology we employ, using LLMs as density estimators for mutual
information measurement, represents a practical approach given the severe
challenges of high-dimensional estimation in long sequences. Alternative methods
like K-NN and neural estimators face fundamental difficulties with
dimensionality and sequence length. While our approach yields consistent
power-law behavior across different models and estimators, both methods likely
underestimate the true exponent, and developing more accurate estimation
techniques remains an important challenge.

Our evaluations rely primarily on open-source models; further verification using
state-of-the-art closed-source models would provide additional validation.

\section*{Broader Impact}

This work advances our theoretical understanding of how language models process
long-range dependencies, with implications for the design and deployment of more
efficient LLM architectures. By establishing the \LsquaredM{} condition, we
provide a principled framework for evaluating an architecture's fundamental
capacity for long-context modeling. This could lead to more efficient models
that maintain effectiveness while reducing computational resources, potentially
decreasing the environmental footprint of training and inference. Our findings
may influence the development of specialized architectures for tasks requiring
long-context understanding, such as legal document analysis, scientific
research, and complex reasoning. 

However, improved long-context modeling could also amplify existing challenges
in LLMs, including the propagation of bias over longer contexts and enhanced
capabilities for generating persuasive misinformation. Research applying the
\LsquaredM{} framework should consider these ethical dimensions, particularly
how improvements in long-range dependency modeling might affect model safety,
fairness, and the verifiability of model outputs.

\section*{Acknowledgements} The authors acknowledge support from the National
Science Foundation under Cooperative Agreement PHY-2019786
(\href{http://iaifi.org/}{The NSF AI Institute for Artificial Intelligence and
Fundamental Interactions}). Z.C. acknowledges support from the Mathworks
Fellowship. Z.C. thanks Rumen Dangovski for helpful discussions. Z.C. and O.M.
thank Amazon Web Services account team, including Brian McCarthy and Jared
Novotny, for technical support. The research was sponsored by the United States
Air Force Research Laboratory and the Department of the Air Force Artificial
Intelligence Accelerator and was accomplished under Cooperative Agreement Number
FA8750-19-2-1000. The computations in this paper were partly run on the FASRC
cluster supported by the FAS Division of Science Research Computing Group at
Harvard University. This research used the DeltaAI advanced computing and data
resource, which is supported by the National Science Foundation (award OAC
2320345) and the State of Illinois, through allocation CIS240904 from the
Advanced Cyberinfrastructure Coordination Ecosystem: Services \& Support
(ACCESS) program, supported by National Science Foundation grants \#2138259,
\#2138286, \#2138307, \#2137603, and \#2138296, and through the National
Artificial Intelligence Research Resource (NAIRR) Pilot NAIRR250043.

\bibliography{reference}
\bibliographystyle{naturemag}

\newpage
\appendix
\onecolumn
\renewcommand{\thesubsection}{\thesection.\Roman{subsection}}
\renewcommand\theequation{\thesection.\arabic{equation}}
\setcounter{equation}{0}
\renewcommand\thefigure{\thesection.\arabic{figure}}
\setcounter{figure}{0}
\renewcommand\thetable{\thesection.\arabic{table}}
\setcounter{table}{0}

\section{Notations and Conventions} \label{app:notation}

{\bf Basic notations}
\begin{itemize}
    \item Random variables: Uppercase letters (e.g., $X$, $Y$, $W$) denote
    random variables.
    \item Realizations: Lowercase letters (e.g., $x$, $y$, $w$) denote specific
    values or realizations of random variables.
    \item Sequences: $X_{i:j}$ denotes the sequence $(X_i, X_{i+1}, \ldots,
    X_j)$.
    \item Bold notation: $\mathbf{X} := X_{i:j}$ when indices are clear from
    context.
    \item Single variable shorthand: $X := X_i$ when the index is clear from
    context.
    \item Logarithms: While the choice of base does not affect the scaling laws
    (only the multiplicative constants), all logarithms are natural logarithms
    (base $e$) unless otherwise specified.
\end{itemize}

{\bf Information-theoretic quantities}
\begin{itemize}
    \item $H(\cdot)$: Shannon entropy.
    \item $H(\cdot|\cdot)$: Conditional (Shannon) entropy.
    \item $I(\cdot;\cdot)$: (Shannon) mutual information.
    \item $D_{\text{KL}}(\cdot||\cdot)$: Kullback--Leibler divergence.
    \item $H^p(\cdot)$: Entropy computed with respect to distribution $p$.
    \item $H^q(\cdot)$: Entropy computed with respect to distribution $q$.
    \item $H(p, q)$: Cross-entropy between distributions $p$ and $q$.
\end{itemize}

{\bf Asymptotic notations}
\begin{itemize}
    \item $f(n) \sim g(n)$: $f$ and $g$ have the same asymptotic growth rate,
    i.e., $f(n) = \Theta(g(n))$.
    \item $f(n) \succ g(n)$: $f$ grows strictly faster than $g$ asymptotically,
    i.e., $f(n) = \omega(g(n))$.
    \item $f(n) \succsim g(n)$: $f$ grows at least as fast as $g$.
    asymptotically, i.e., $f(n) = \Omega(g(n))$.
    \item $f(n) \prec g(n)$: $f$ grows strictly slower than $g$ asymptotically,
    i.e., $f(n) = o(g(n))$.
    \item $f(n) \precsim g(n)$: $f$ grows at most as fast as $g$ asymptotically,
    i.e., $f(n) = O(g(n))$.
\end{itemize}

{\bf Distributions and expectations}
\begin{itemize}
    \item $p$: True underlying probability distribution (of natural language).
    \item $q$: Model-generated probability distribution (sometimes refers to the
    model itself).
    \item $\mathbb{E}_p[\cdot]$: Expectation with respect to distribution $p$.
    \item $p_X \otimes p_Y$: Product distribution of marginals $p_X$ and $p_Y$.
\end{itemize}

{\bf Model-specific notations}
\begin{itemize}
    \item $w_{\text{BOS}}$: Beginning-of-sequence token.
    \item $M$: Vocabulary size.
    \item $L$: Sequence length.
    \item $\ell$: Position of sequence split for bipartite mutual information.
    \item $\dim(\bz)$: Dimensionality of the history state $\bz$.
    \item $\theta$: Model parameters.
\end{itemize}

{\bf Special notations}
\begin{itemize}
    \item $I^{\text{BP}}$: Bipartite mutual information.
    \item $I^{\text{TP}}$: Two-point mutual information.
    \item $\mathbbm{1}(\cdot)$: Indicator function (equals 1 when condition is
    true, 0 otherwise).
\end{itemize}

The notations are used consistently throughout the main text and appendices
unless otherwise specified in local contexts.

\section{Additional Details on Mutual Information Scalings}
\label{app:add_mutual_info_scale}

\subsection{Bipartite Mutual Information Scaling with Additional LLMs on
Additional Datasets} \label{app:mi_other_llm}

In the main text, we use the LLaMA 3.1 405B model as the density estimator and
measured the bipartite mutual information scaling on \PG{} dataset. In this
section, we provide additional estimations of the bipartite mutual information
scaling using the DeepSeek V3 Base model and on \Wikipedia{} dataset. We note
that because we are merely measuring the conditional probabilities of the input
tokens without interactions with the agent, we believe the
non-instruction-finetuned model better suits our tasks.

\begin{figure}[ht]
    \centering
    \includegraphics[width=0.67\linewidth]{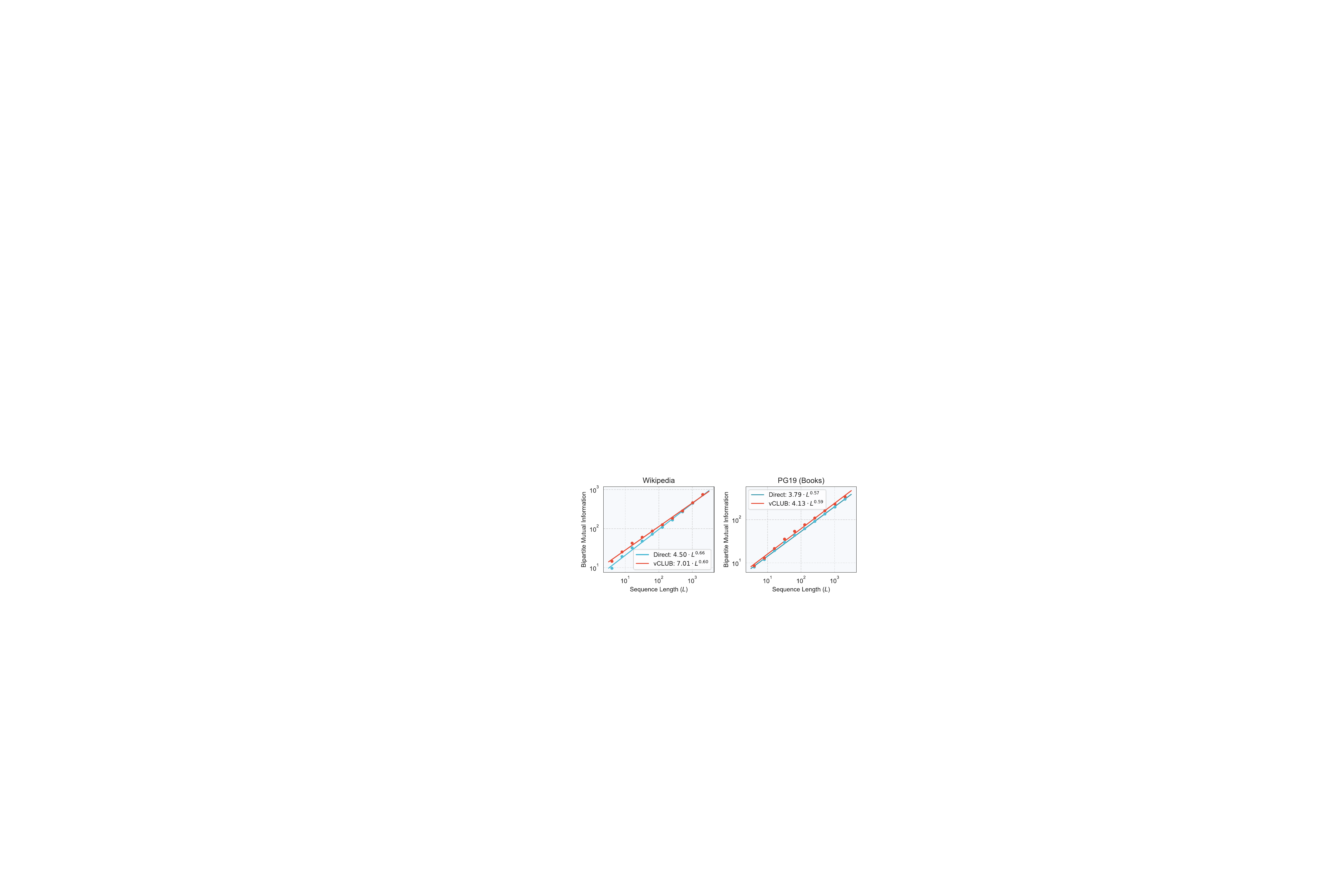}
    \caption{Bipartite mutual information estimation using (left) LLaMA 3.1 405B
    on the \Wikipedia{} dataset and (right) Deepseek V3 Base model on the \PG{}
    dataset. All direct measurements include the bias correction described in
    Appx.~\ref{app:mi_direct}.}
    \label{fig:mi_different_model}
\end{figure}

In Fig.~\ref{fig:mi_different_model}, we report the results on \Wikipedia{}
dataset using LLaMA 3.1 405B model as the density estimator and on \PG{} dataset
using Deepseek V3 Base model as the density estimator. We find that in both
cases, clear sub-volume growth behavior is observed. We note that the measured
exponent should be taken with a grain of salt and likely underestimates the true
mutual information scaling due to reasons explained in
Appx.~\ref{app:underestimate}.

\subsection{Bipartite Mutual Information Scaling Under Various Ratios of $\ell /
L$} \label{app:mi_other_l}

In the main text, we focused on the bipartite mutual information with equal
splits. However, the bipartite mutual information scaling is not limited to
equal biparitition. In this section, we provide additional results for various
ratios of $\ell/L$. 

\begin{figure}[ht]
    \centering
    \includegraphics[width=0.67\linewidth]{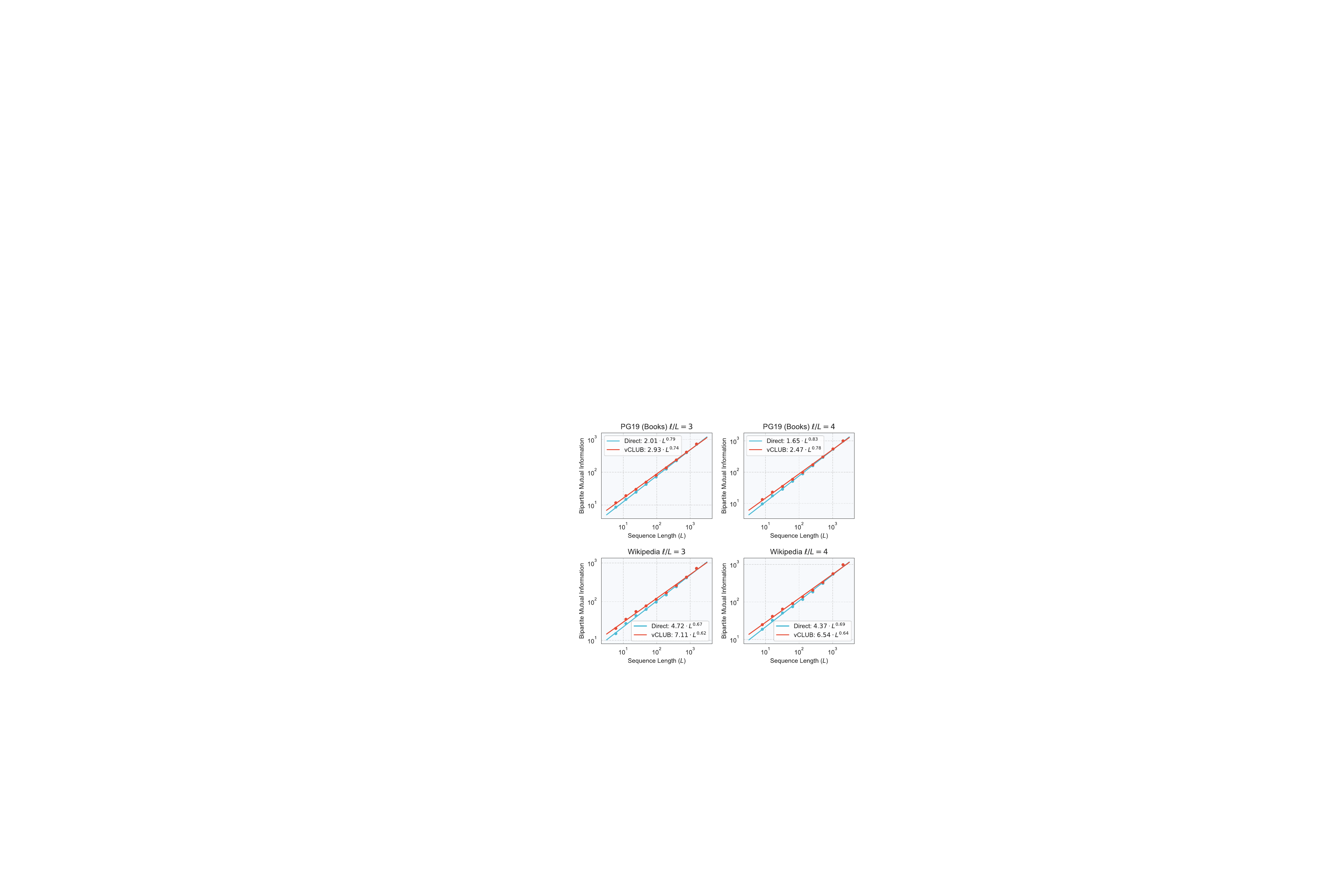}
    \caption{Bipartite mutual information estimation using different ratios of $\ell/L$. All results suggest the existence of power-law scaling, with various fitted exponents.}
    \label{fig:mi_different_ratio}
\end{figure}

In Fig.~\ref{fig:mi_different_ratio}, we provide estimation of the bipartite
mutual information scaling for $\ell/L=3$ and $\ell/L=4$. All results show clear
power-law relations, and are consistent with
Fig.~\ref{fig:real_data_measurement} in the main text. These results can be used
to support the \LsquaredM{} condition with similar arguments as in the main
text.

\subsection{Why The Estimated Exponent $\beta$ Is Likely An Underestimation?}
\label{app:underestimate} In the main text, we mentioned that our measured
exponent $\beta$ using LLMs likely underestimates the true $\beta$. Here, we
discuss the reasons.

For the direct estimator,
\begin{equation}
    I_{\ell;L}^{\mathrm{BP},\mathrm{direct}} =H(p_{\bY}, q_{\bY}) -  H(p_{\bY|\bX}, q_{\bY|\bX}),
\end{equation}
both terms (without the minus sign) overestimates the true (conditional)
entropy, but for different extent and at different scales. 

At small $L$, the first term suffers from the bias from the BOS token as
discussed in Appx.~\ref{app:mi_direct}. The second term, despite also an
overestimation, does not suffer from the BOS token issue. Therefore, at small
$L$, the direct estimator tends to overestimate the true entropy. 

At large $L$, the bias from the BOS token is less severe. However, modeling
$p(\bY|\bX)$ requires the model to correctly capture all the dependencies
between $\bX$ and $\bY$, making it significantly harder than modeling $p(\bY)$
alone. Therefore, $q(\bY|\bX)$ is likely a worse estimation of the true
distribution than $q(\bY)$, resulting in more overestimation in the second term,
and an underestimation of the bipartite mutual information. 

This means that the direct estimator tends to overestimate the true bipartite
mutual information at small $L$ and underestimate it at large $L$, resulting in
an underestimation of the fitted exponent.

The vCLUB estimator, as pointed out in \cite{vclub}, is an upper bound to the
true mutual information if $q$ is close to $p$, but fails to maintain the
property when the KL-divergence between them increases. Therefore, it is likely
that this estimator also overestimates the true bipartite mutual information at
small $L$ and underestimates it at large $L$, resulting in a similar
underestimation of the fitted exponent as our direct estimator. As our fitted
exponent for the vCLUB estimator is smaller than that of the direct estimator,
we conclude that the vCLUB estimator has a larger bias in this case, and it is
reasonable to believe that the true exponent is even larger.

\subsection{Direct Estimation of Bipartite Mutual Information Using LLMs}
\label{app:mi_direct} In the main text, our direct estimator for the bipartite
mutual information is
\begin{equation}
    I_{\ell;L}^{\mathrm{BP},\mathrm{direct}} =\expect_{p_{\bX\bY}} \left[ \log q(\bY|\bX) - \log q(\bY) \right]
    = H(p_{\bY}, q_{\bY}) -  H(p_{\bY|\bX}, q_{\bY|\bX})
    = I^{p}(\bX;\bY) + \varepsilon(p, q).
\end{equation}
where as usual, $\bX:=X_{1:\ell} := W_{1:\ell}$ and $\bY:=Y_{1:L-\ell} :=
W_{\ell+1:L}$ with $W_{1:L}$ being a sequence of tokens. However, as discussed
in the main text, the $H(p_{\bY}, q_{\bY})$ term suffers from an additional
bias---we cannot guarantee that $\bY$ starts at the beginning of a sentence, but
LLMs model distributions conditioned on BOS token. To mitigate this issue, we
use $n$-gram calculations to correct the entropy of the first two tokens as
explained below.

We first rewrite the (marginal) cross entropy as
\begin{equation}
    H(p_{\bY}, q_{\bY}) = -\expect_p [\log q(\bY)] = -\sum_{i=1}^{L-\ell}\expect_p [\log q(Y_i|Y_{1:i-1})], 
\end{equation}
where as usual, the expectation over the conditional variable is omitted but
implied in the cross entropy calculation.

In modern LLMs, we can only compute $q(y_i|y_{1:i-1},w_{BOS}) \ne
q(y_i|y_{1:i-1})$, resulting in an additional error in the bipartite mutual
information estimation. In practice, this difference becomes less pronounced for
larger $i$, because it matters less if the sequence starts at the beginning of a
sequence or not if there are many $y_{1:i-1}$ prior tokens to conditional on.
Therefore, we focusing on reducing the bias for small $i$. In addition, if $i$
is small, we can iterate over the dataset and construct a histogram for the
$i$-gram distribution $p(y_{1:i})$. 

We denote the count for each $i$-tuple of tokens with $n_{y_{1:i}}$ and the
total number of samples with $N$. Then, the entropy of the distribution can be
estimated naively as 
\begin{equation}
    \hat{H}^{\text{na\"ive}}(Y_{1:i}) = - \sum_{y_{1:i}} \frac{n_{y_{1:i}}}{N}\log \frac{n_{y_{1:i}}}{N} = \log N - \frac{1}{N}\sum_{y_{1:i}} n_{y_{1:i}} \log n_{y_{1:i}},
\end{equation}
where the summation runs over all possible combination of tokens
$y_{1:i}:=(y_1,y_2,\dots y_i)$. 

However, this estimation is severely biased and underestimates the true entropy,
due to the concavity of logarithm function. In \cite{entropybiascorrect}, a
bias-corrected estimator is proposed by replacing the logarithm function with a
new function 
\begin{equation} \label{eq:h_G_app}
    \hat{H}^{G}(Y_{1:i}) = \log N - \frac{1}{N}\sum_{y_{1:i}} n_{y_{1:i}} G(n_{y_{1:i}}),
\end{equation}
where 
\begin{equation}
    G(n) = \psi(n) + \frac{(-1)^n}{2}\left(\psi\left(\frac{n+1}{2}\right) - \psi\left(\frac{n}{2}\right)\right),
\end{equation}
with $\psi(\cdot)$ the digamma function. We note that
Ref.~\cite{entropybiascorrect} was not able to obtain the closed form expression
for $G(\cdot)$, which we derived with the help of Wolfram Mathematica
\cite{Mathematica}.

\begin{figure}[ht]
    \centering
    \includegraphics[width=0.67\linewidth]{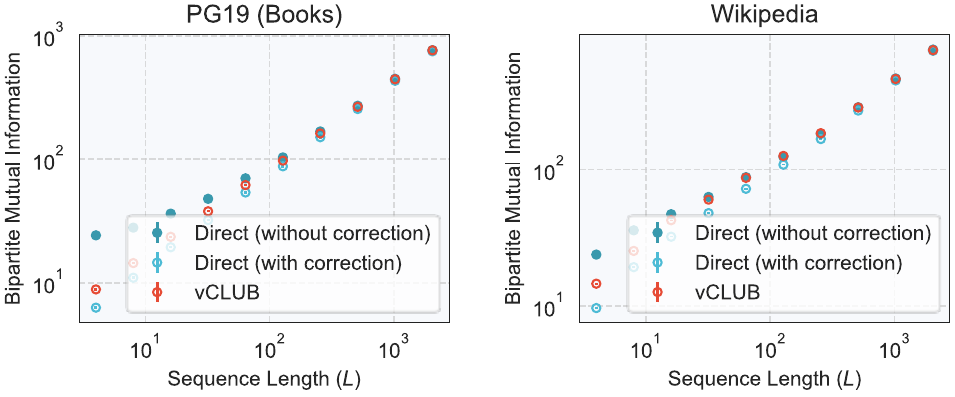}
    \caption{Effect of bias correction method in the direct estimator. The bias only affects the estimation at small sequence lengths, and all methods converge at large sequence lengths.}
    \label{fig:mi_correction_effect}
\end{figure}

This bias-corrected estimator still underestimates the true entropy, but much
less compared to the original na\"ive estimator. In the main text, we estimate
the (marginal) cross entropy with $2$-gram correction in the following way.
Breaking up the cross entropy as
\begin{equation}
    H(p_{\bY}, q_{\bY}) = -\sum_{i=3}^{L-\ell}\expect_p [\log q(Y_i|Y_{1:i-1})] + H(p_{Y_1 Y_2}, q_{Y_1 Y_2}).
\end{equation}
For the first term, we use LLM generated $q(y_i|y_{1:i-1},w_{BOS})$ as
approximation. For the second term, we mitigate the bias from LLM estimation by
combining it with Eq.~\eqref{eq:h_G_app} as $H(p_{Y_1 Y_2}, q_{Y_1
Y_2|w_{BOS}})/5 + 4 \hat{H}_p^G(Y_1 Y_2)/5$. In
Fig.~\ref{fig:mi_correction_effect}, we also present the result without this
correction and show that this bias correction mostly affects the estimation at
small lengths $L$, and does not alter the general scaling behavior. In addition,
since the result from this bias-corrected direct estimator agrees with the vCLUB
\cite{vclub} estimator, we believe this correction is reasonable.

\subsection{Additional Discussion on Mutual Information Estimation Methods}
\label{app:mi_estimation_methods}

In the main text, we briefly discussed the limitations of traditional mutual
information estimation methods for our high-dimensional, long-sequence setting.
Here we provide additional technical details on why these methods are
challenging to apply to our settings.

\textbf{Neural Estimators: MINE and InfoNCE.}
Neural estimators like MINE \citep{belghazi2021minemutualinformationneural} and
InfoNCE \citep{oord2019representationlearningcontrastivepredictive} train deep
neural networks as critics to estimate mutual information. Both methods can
fundamentally be viewed as training unnormalized density estimators or density
ratio estimators.

MINE uses the Donsker--Varadhan representation of KL divergence
\citep{donsker1975asymptotic} and trains a critic $T_\theta(x,y)$ to maximize:
\begin{equation}
\mathbb{E}_{p(x,y)}[T_\theta(x,y)] - \log \mathbb{E}_{p(x)p(y)}[e^{T_\theta(x,y)}]
\end{equation}
The optimal critic approximates the log density ratio $\log
\frac{p(x,y)}{p(x)p(y)}$. However, this objective suffers from high variance and
numerical instability when mutual information is large, which is especially
challenging given the high-dimensional nature of long sequences we analyze.

InfoNCE uses noise-contrastive estimation with multiple negative samples:
\begin{equation}
I(X;Y) \geq \mathbb{E}\left[\log \frac{e^{f(x_i,y_i)}}{\frac{1}{K}\sum_{j=1}^K e^{f(x_i,y_j)}}\right]
\end{equation}
where the expectation is over $K$ independent samples from the joint
distribution. The bound is upper bounded by $\log K$, which means for our
setting where mutual information can be on the order of thousands, this would
require prohibitively large batch sizes to obtain accurate estimates.

Both methods require training critics from scratch to learn representations of
natural language distributions, which could require datasets and computational
resources comparable to training LLMs themselves.

Other variational bounds \citep{poole2019variationalboundsmutualinformation}
face similar challenges.

\textbf{K-Nearest Neighbor Estimators.}
K-nearest neighbor (K-NN) estimators
\citep{gao2015efficientestimationmutualinformation}
estimate mutual information based on distances between samples in joint and
marginal spaces. While asymptotically unbiased and training-free, they also face
challenges for text.

Text consists of discrete tokens that must be embedded into continuous spaces
for K-NN estimation. Modern token embeddings have dimensions in the thousands,
and for sequences of thousands of tokens, the combined dimensionality can make
K-NN estimation impractical as the number of samples required for reliable K-NN
estimates grows exponentially with dimension.

\textbf{Connection to Our LLM-Based Approach.}
Our approach leverages pre-trained LLMs as density estimators, providing
$q(y|x)$ directly through conditional probabilities and approximating $q(y)$
efficiently. This avoids training critics from scratch and the curse of
dimensionality from distance-based estimation. We believe this is well-suited
for our use case of analyzing long natural language sequences.

\subsection{Estimation of Two-Point Mutual Information} \label{app:two-point} In
the main text, we included the results for two-point mutual information for
completeness. In this section, we explain how the results are obtained. 

Two-point mutual information estimation is more straightforward compared to
bipartite mutual information, requiring only 1- and 2-gram statistics without
LLM approximations. We estimate this quantity using entropy calculations for
individual tokens and token pairs separated by distance $d$. Following
\cite{entropybiascorrect}, we employ their bias-reduced entropy estimator:
\begin{equation}
    \hat{H}^{G}(X)=\hat{H}^{G}(Y) = \log N - \frac{1}{N} \sum_{m=1}^M n_m G(n_m),
\end{equation}
where $N$ is the total number of tokens, $M$ is the vocabulary size, $n_m$ is
the number of tokens whose token ID equals $m$, and $G(\cdot)$ is defined as
\begin{equation}
    G(n) = \psi(n) + \frac{(-1)^{n}}{2} \left(\psi\left(\frac{n+1}{2}\right) - \psi\left(\frac{n}{2}\right)\right)
\end{equation}
with $\psi(\cdot)$ the digamma function.

The entropy of pairs of tokens is estimated analogously, with the summation
running over all ordered pairs of tokens $(m_i, m_j)$, resulting in the total
number of terms quadratic in the vocabulary size. The mutual information is then
estimated as

\begin{equation}
    \hat{I}_d^{\mathrm{TP}}(X;Y)=\hat{H}^{G}(X) + \hat{H}^{G}(Y) - \hat{H}^{G}(XY).
\end{equation}

We note that this mutual information estimator exhibits systematic bias. The
entropy estimator has a negative bias that increases (in absolute value) with
dimension of the sample space $\abs{\Omega}$. Since
$\abs{\Omega_{X}}=\abs{\Omega_{Y}}=M$ where as $\abs{\Omega_{XY}}=M^2$, the bias
in $\hat{H}(XY)$ exceeds that in $\hat{H}(X)=\hat{H}(Y)$, leading to a positive
bias in $\hat{I}_d$. This bias becomes particularly problematic at large
distances $d$, where $H(XY) \approx H(X) + H(Y)$ and $I_d$ approaches zero. To
mitigate this issue, we perform additional bias correction by fitting the
systematic bias from the data (see Appx.~\ref{app:two-point-bias-correction} for
details).

\begin{figure}[ht]
    \centering
    \includegraphics[width=0.67\linewidth]{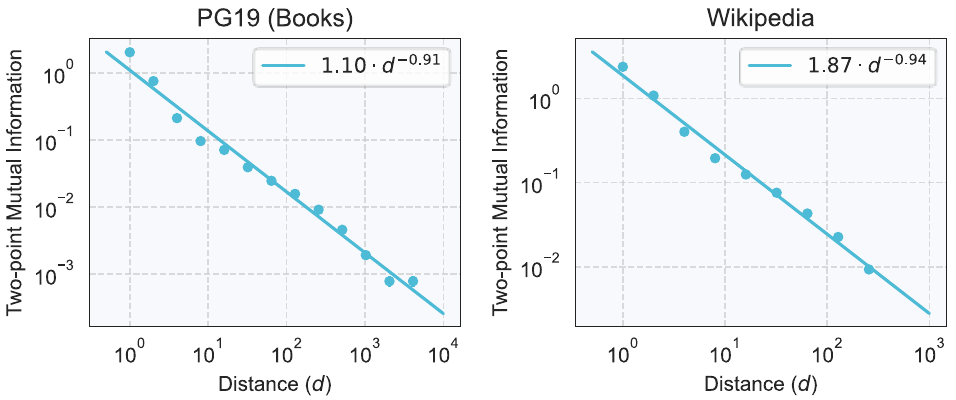}
    \caption{Two-point mutual information scaling on \PG{} and \Wikipedia{}
    datasets.}
    \label{fig:two_point_mi}
\end{figure}

We apply this methodology to measure two-point mutual information on both the
\PG{} dataset and \Wikipedia{}, confirming power-law decay in both cases
[Fig.~\ref{fig:two_point_mi}].

\subsection{Bias Correction for Two-point Mutual Information}
\label{app:two-point-bias-correction}

As discussed previously, the estimation of two-point mutual information can be
calculated directly using the $n$-gram approximation [Eq.~\eqref{eq:h_G_app}],
and compute the two-point mutual information as
\begin{equation}
    \hat{I}_d^{\mathrm{TP}}(X;Y)=\hat{H}^{G}(X) + \hat{H}^{G}(Y) - \hat{H}^{G}(XY).
\end{equation}
As discussed in Appx.~\ref{app:mi_direct}, this entropy estimator has a negative
bias, whose magnitude depends on the ratio $\abs{\Omega}/N$, with $\abs{\Omega}$
the size of the corresponding sample space. Since the sample space for the joint
distribution is larger, it has a larger negative bias, resulting in a positive
bias in $\hat{I}$. When $d$ is small, this bias is relatively small compared to
the mutual information itself. However, as $d$ becomes larger, $X$ and $Y$
become less correlated, and $H(XY)\rightarrow H(X)+H(Y)$. In this case, the
estimator can be dominated by this bias, and fitting for the power-law exponent
becomes impossible.

To mitigate this issue, we propose a bias-corrected estimator.

\begin{equation}
    \hat{I}_d^{\mathrm{TP},\mathrm{corrected}}(X;Y)=\hat{H}^{G}(X) + \hat{H}^{G}(Y) - \hat{H}^{G}(XY) - C,
\end{equation}
where $C$ is an unknown positive constant that does not depend on the distance
$d$, which accounts for the bias of the original estimator.

To obtain this bias correction term and fit the power-law exponent, we minimize
the following loss function
\begin{equation}
    \sum_d(\log{}(\hat{I}_d^{\mathrm{TP}}-C) - (\log A - \alpha \log d))^2,
\end{equation}
which is just $\hat{I}_d^{\mathrm{TP}} = A d^{-\alpha} + C$ fitted in log-log
space. Then, we take the fitted $C$ as the systematic bias and $\alpha$ as the
fitted power-law exponent.

\begin{figure}[ht]
    \centering
    \includegraphics[width=0.67\linewidth]{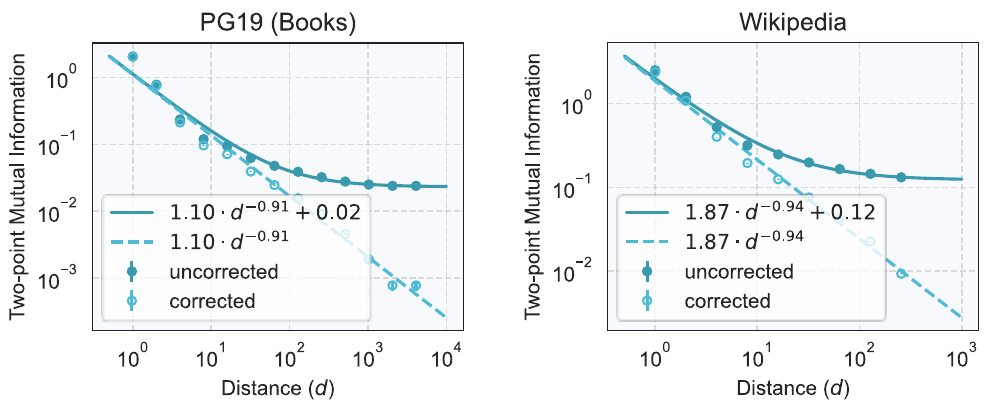}
    \caption{Effect of bias correction for two-point mutual information. The bias causes a plateau at large distances.}
    \label{fig:two_point_mi_correction_effect}
\end{figure}

In Fig.~\ref{fig:two_point_mi_correction_effect}, we compare the corrected and
uncorrected two-point mutual information as a function of $d$ (only the
corrected version is shown in the main text). Without the bias correction, the
data appear to have larger long-range dependencies, but after the bias
correction, all points lie on a straight line in a log-log plot. The bias
correction constant is much smaller than the entropies involved in the
calculation, even the smallest two-token entropy measured is $12.5$, at least
two orders of magnitude larger than the fitted bias correction. In addition, the
fact that a single variable added to the fitting function can fit the data so
well suggests the bias correction is reasonable and highly effective.

We note that on \Wikipedia{}, we were only able to measure the two-point mutual
information up to $d=256$, due to limited long-context length data in
\Wikipedia{}.

\subsection{Failures of Two-point Mutual Information}
\label{app:two-point-failure}

In this section, we further demonstrate the relation between two-point and
bipartite mutual information, and why two-point mutual information cannot
properly capture the full multi-token dependencies needed for modeling natural
language. 

In existing literature, the scaling of two-point mutual information has been
used to demonstrate the existence of long-range dependencies in natural
language. In particular, it is believed that short-range dependencies are
characterized by an exponential decay in two-point mutual information, as seen
typically in finite-state Markov chains, and the existence of power-law decay in
two-point mutual information in natural language indicates non-trivial
long-range dependencies. Although this perspective is correct if one defines the
existence of long-range dependence as as the existence of non-exponential-decay
two-point mutual information, this definition does not properly account for the
mutual information between token pairs when other tokens are present. This can
be made more clear by considering the following decomposition of bipartite
mutual information.

For a sequence of tokens $W_{1:L}$ with $X_{1:\ell} = W_{1:\ell}$ and
$Y_{1:L-\ell} = W_{\ell+1:L}$, the bipartite mutual information reads 
\begin{equation}
    I^{\mathrm{BP}}_{\ell;L} = I(X_{1:\ell}; Y_{1:L-\ell}).
\end{equation}
Standard information theory allows mutual information to be decomposed as
\begin{equation}
    I(XZ;Y) = I(X;Y) + I(Z;Y|X),
\end{equation}
where $I(Z;Y|X)$ is the conditional mutual information between $Z$ and $Y$ given
$X$. Using this relation repeatedly, the bipartite mutual information can be
decomposed as
\begin{equation}
\begin{aligned}
    I^{\mathrm{BP}}_{\ell;L} &= I(X_{1:\ell}; Y_{1:L-\ell}) \\
    &= I(X_1;Y_1) + I(X_2;Y_1|X_1) + I(X_1;Y_2|Y_1) + I(X_2;Y_2|X_1Y_1) + \cdots \\
    &= \sum_{i=1}^{\ell} \sum_{j=1}^{L-\ell}I(X_i;Y_j|X_{1:i-1}Y_{1:j-1}) \\
    &\ne \sum_{i=1}^{\ell} \sum_{j=1}^{L-\ell} I(X_i;Y_j) = \sum_{i=1}^{\ell} \sum_{j=1}^{L-\ell} I^{\mathrm{TP}}_{j-i+\ell}.
\end{aligned}
\end{equation}

In fact, it is in general not even clear whether the conditional mutual
information $I(X_i;Y_j|X_{1:i-1}Y_{1:j-1})$ is greater or less than the marginal
mutual information $I(X_i;Y_j)$. Therefore, as demonstrated here, when
considering dependencies between blocks of text, a simple aggregation of the
two-point mutual information gives a very incomplete picture. Due to this
reason, weakly correlated systems, such as the example mentioned in
Sec.~\ref{sec:distiction} could exhibit seeming strong long-range two-point
dependencies, and systems with very different bipartite mutual information,
could share very similar two-point information scaling, as we will show later in
Appx.~\ref{app:gaussian}.

\section{Multivariate Gaussian Distributions} \label{app:gaussian} In the main
text, we considered two families of multivariate Gaussian distributions of
different sequence lengths to demonstrate the distinction between bipartite and
two-point mutual information scalings. In particular, one is designed to mimic
natural language, both in terms of the sub-volume law growth of the bipartite
mutual information and the power-law decay of two-point mutual information. This
family of distributions is also used to empirically verify our theory on
\LsquaredM{} condition for different LLM architectures. The other is designed to
have the same two-point mutual information scaling, but very different bipartite
mutual information scaling, showcasing that one can have distributions with the
same two-point mutual information scaling, but drastically different bipartite
mutual information scalings.

\subsection{Mutual Information Scalings}
\begin{figure}[ht]
    \centering
    \includegraphics[width=0.67\linewidth]{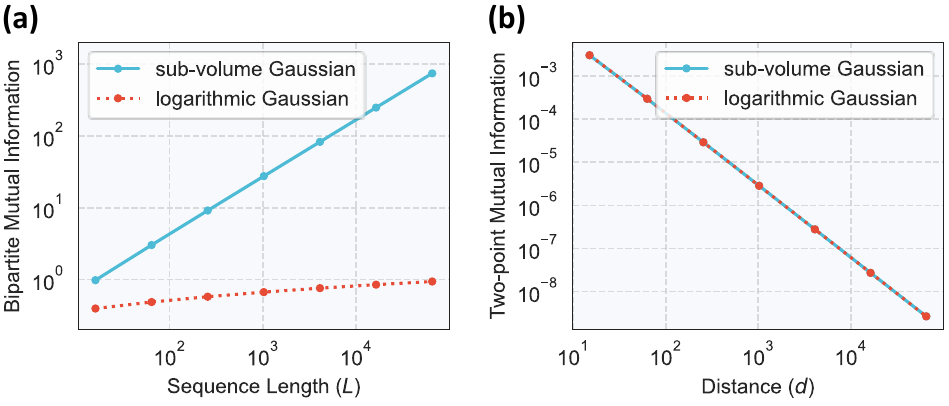}
    \caption{Bipartite and two-point mutual information of the two families of Gaussian distributions.}
    \label{fig:syn_data_measurement}
\end{figure}
Before showing the construction details, we first present both the bipartite and
two-point mutual information scalings of the two families of Gaussian
distributions. As shown in Fig.~\ref{fig:syn_data_measurement}, they have
drastically different bipartite mutual information scaling---one has a power-law
relation and the other logarithmic---but their antipodal two-point mutual
information scaling is identical. This further demonstrates that two-point
mutual information alone gives incomplete information of the multi-token
long-range dependencies present in sequence data, and the simple aggregation of
two-point mutual information does not tell the full picture of the bipartite
mutual information. In fact, one can construct distributions with the same
power-law decay in two-point mutual information, but has a constant bipartite
mutual information scaling as well.

\subsection{Construction} \label{app:gaussian_construction} Let's start by
considering the family of distributions with sub-volume law growth. The
distributions are constructed in a hierarchical manner. 

We start at the first layer, with four independent standard Gaussian random
variables $(X_1,X_2,X_3,X_4)$. Then, define the change-of-coordinate matrix
\begin{equation}
    \mathcal{M} = \mqty(\gamma & \gamma& \gamma& \rho \\
                        -\gamma & \gamma & -\gamma & \rho \\
                        -\gamma & -\gamma & \gamma & \rho \\
                        \gamma & -\gamma & -\gamma & \rho),
\end{equation}
where we choose $\gamma = \sqrt{5}/4$ and $\rho = 1/4$. The output of the first
layer is defined as
\begin{equation}
    \bY = \mathcal{M} \bX,
\end{equation}
where the random variables are now correlated. It is easy to verify that this
operation only changes the off-diagonal elements in the covariance matrix, and
leaves the diagonal elements unaffected.

For the second layer and up, we first stack three independently sampled copies
from the previous layer and attach additional independent standard Gaussian
random variables as the fourth elements as
\begin{equation} \label{eq:concat_app}
    \mathcal{X} = \mqty(Y_{1,1} & Y_{1,2} & Y_{1,3} & W_{1}\\
          Y_{2,1} & Y_{2,2} & Y_{2,3} & W_{2} \\
          Y_{3,1} & Y_{3,2} & Y_{3,3} & W_{3} \\
          \vdots & \vdots & \vdots & \vdots),
\end{equation}
where $Y_{i,j}$ refers to the $i$th output from previous layer of the $j$th
copy, and $W_{i}$ refers to the $i$th newly sampled standard Gaussian random
variable.

Note that at this point, all rows are independent from each other; therefore we
apply the change of coordinate matrix $\mathcal{M}$ at each row to correlate
them. The matrix is then flattened to obtain $\bZ$
\begin{equation}
    \bZ = (Z_{1,1}, Z_{1,2}, Z_{1,3}, Z_{1,4}, Z_{2,1}, Z_{2,2}, Z_{2,3}, Z_{2,4}, Z_{3,1}, Z_{3,2}, Z_{3,3}, Z_{3,4}, \cdots),
\end{equation}
where the subscripts denote the variables' original position in the matrix.
Before outputting from this layer, we perform an addition operation to each pair
of random variables $(Z_{i,4}, Z_{i+1,1})$, by applying a coordinate
transformation that modifies their correlations as
\begin{equation}
    \mathrm{corr}(Z_{i,4}, Z_{i+1,1}) \rightarrow \frac{2}{5}(\mathrm{corr}(Z_{i,3}, Z_{i,4}) + \mathrm{corr}(Z_{i+1,1}, Z_{i+1,2})) + \frac{1}{5}.
\end{equation}
This operation may seem arbitrary, but it is crucial to introduce correlations
that give a linear ordering of the random variables. Without this operation, the
distribution simply forms a tree structure.

Now, we can truncate the construction at different layers $l$ and form a family
of distributions with different sequence lengths $L=4^l$. In
Fig.~\ref{fig:syn_data_measurement} of the main text, we consider up to $8$
layers, and in Fig.~\ref{fig:syn_kl_div} and~\ref{fig:real_nll}, we consider
$l=4$, $5$ and, $6$.

The second family of distributions is constructed analogously. The only
difference is that we replace Eq.~\eqref{eq:concat_app} with a single copy of
$\bY$ and three independent copies of $\bW$ as

\begin{equation}
    \mathcal{X} = \mqty(Y_1 & W_{1,1} & W_{1,2} & W_{1,3}\\
          Y_2 & W_{2,1} & W_{2,2} & W_{2,3} \\
          Y_3 & W_{3,1} & W_{3,2} & W_{3,3} \\
          \vdots & \vdots & \vdots & \vdots),
\end{equation}

\subsection{Properties} \label{app:syn_dist_prop} These two series of
distributions have many nice properties, in addition to their bipartite and
two-point mutual information scalings. In addition, these constructions directly
defines the multi-variate probability distribution due to their Gaussian nature.
This allows for exact calculations of conditional probability distributions for
training LLMs, as well as direct computation of the bipartite and two-point
mutual information without LLM approximations.

\section{Model State for Storing Past Information} \label{app:history_state} In
Definition~\ref{def:model_state_z} in the main text, we give a concrete
definition of ``model state for storing past information'' as history state, and
claim that it is the past key-value pairs for transformers and recurrent state
for SSMs and RNNs. Here, we explain them in more detail.
\subsection{Transformers}
In transformers, only the attention block mixes information among different
tokens, therefore we only need to analyze the behavior of the attention block.
We will be assuming the existence of the causal mask, as our theory mainly
applied to autoregressive LLMs. Denoting the input and output of the attention
layer as $\bx$ and $\by$ (notice they are no longer two parts of a sequence),
the self-attention mechanism is defined as
\begin{equation}
    \boldsymbol{y} = \mathrm{softmax}((W_q \bx) (W_k \bx)^T) W_v \bx,
\end{equation}
where $W_q$, $W_k$ and $W_v$ are the weight matrices. For simplicity, we drop
the usual $\sqrt{h_{\mathrm{dim}}}$ normalization and the output weight matrix,
as they are irrelevant to our discussion.

Separating the calculation for each token, the mechanism can be rewritten as
\begin{equation} \label{eq:kv_cache_app}
    y_i =  \frac{\sum_{j=1}^{i} e^{(W_q x_i) (W_k x_j)}  W_v x_j}{\sum_{j'=1}^{i} e^{(W_q x_i) (W_k x_{j'})}} = \frac{e^{(W_q x_i) (W_k x_i)} W_v x_i + \sum_{j=1}^{i-1} e^{(W_q x_i) k_j}  v_j }{e^{(W_q x_i) (W_k x_i)} + \sum_{j'=1}^{i-1} e^{(W_q x_i) k_{j'}}},
\end{equation}
where $k_j = W_k x_j$ and $v_j = W_v x_j$ are keys and values which we sum over
past tokens. Clearly, the attention output only depends on the current token
$x_i$ and the past key-value pairs $k_{1:i-1}$ and $v_{1:i-1}$. This arguments
extends to all $y_k$ with $k\ge i$, where all $y_k$'s dependency on $x_{1:i-1}$
is via $k_{1:i-1}$ and $v_{1:i-1}$. Therefore, key-value pairs form the history
state, and their size grows linearly with input sequence length. We note that
Eq.~\eqref{eq:kv_cache_app} also describes how key-value caching works.
\subsection{State Space Models and RNNs}
State space models (SSMs) and RNNs, on the other hand, are easier to analyze.
These models in general all have some recurrent state with a fixed size, and
some mechanism to update the state when a new token is observed. The output
depends only on the previous recurrent state, and the current token. They can in
general be written in the following way.
\begin{equation}
\begin{aligned}
    h_i &= f(h_{i-1}, x_i), \\
    y_i &= g(h_{i-1}, x_i),
\end{aligned}
\end{equation}
for some update function $f$ and output function $g$. It is obvious that the
history state is exactly this recurrent state (or the collection of recurrent
states of different layers), which does not grow with the input sequence.

We note that this discussion also applies to linear attention models, whose
key-value pairs can be merged into a recurrent state with fixed size, due to the
replacement of softmax function. Test time training (TTT) models can also be
included in this discussion. They can be viewed as RNNs with inner model
parameters as recurrent state, and test time training process as update
function.

\subsection{Other Architectures}
For other models, such as sparse transformers or some compression-based models,
the analysis has to be performed separately. Nevertheless, the \LsquaredM{}
framework is general: after identifying the history state, one can always
compare its scaling with the bipartite mutual information scaling to see whether
the model is capable of capturing the long-range dependencies in the data.

\section{Proofs of Theorem~\ref{thm:mi_to_z}} \label{app:proof}

We provide three proofs of Theorem~\ref{thm:mi_to_z} under different
assumptions, demonstrating the universality and robustness of the result.
Importantly, all three sets of assumptions are extremely mild and directly
reflect realistic conditions in modern neural networks, whether through the
discrete nature of floating-point arithmetic, empirically observed geometric
properties of neural representations, or basic continuity requirements.

\subsection{Proof Under Discreteness Assumption}

The discreteness assumption is already quite reasonable in practice. Modern
neural networks use floating-point representations, which are inherently
discrete with finite precision. Moreover, neural networks have been shown to
retain strong performance even under aggressive quantization, demonstrating that
discrete representations with limited precision are sufficient to capture the
essential information. This discreteness assumption thus provides a natural and
practical starting point for our proof.

\begin{theorem}[Theorem~\ref{thm:mi_to_z}, Discrete Version]
Assume the history state $\bz_\ell$ takes discrete values. Then a model's
capacity to capture bipartite mutual information is bounded by the size of its
history state as
\begin{equation}
I_{\ell;L}^{\mathrm{BP},q} \leq C \cdot \dim(\boldsymbol{z}_{\ell}) + \log(M)
\end{equation}
where $C$ is some constant and $M$ denotes the vocabulary size.
\end{theorem}

\begin{proof}
By the data processing inequality: $I^{q}(X_{1:\ell};Y_{1:L-\ell}) \leq
I^{q}(\boldsymbol{Z}_{\ell} X_{\ell};Y_{1:L-\ell})$. This is upper bounded by
the entropy: $H^{q}(\boldsymbol{Z}_{\ell} X_{\ell})$, which is further upper
bounded by $ H^{q}(\boldsymbol{Z}_{\ell}) + H^{q}(X_{\ell}) \leq C \cdot
\dim(\boldsymbol{z}_{\ell}) + \log(M)$, where the last inequality follows from
the bound on entropies of discrete variables.
\end{proof}

\subsection{Proof Under Almost Orthogonal Directions (AOD) Assumption}

The discreteness assumption can be relaxed if we instead assume the following
observed fact about neural networks: neural networks store distinct information
in almost orthogonal directions (AODs) of the hidden state
\cite{elhage2022toymodelssuperposition,eb44ce1f7e1f4deac10f6e7009e2073f1eb0b3e4,8899a672040d67e8f58d1bc3efcbc966a7798c7d}. 

\begin{theorem}[Theorem~\ref{thm:mi_to_z}, AOD Version]
Assume neural networks store distinct information in almost orthogonal
directions (AODs) of the hidden state. Then a model's capacity to capture
bipartite mutual information is bounded as
\begin{equation}
I_{\ell;L}^{\mathrm{BP},q} \leq C \cdot \dim(\boldsymbol{z}_{\ell}) + \log(M)
\end{equation}
where $C$ is some constant and $M$ denotes the vocabulary size.
\end{theorem}

\begin{proof}
An autoregressive neural network's dependency on past tokens is through the
intermediate variable $\bz_\ell = \boldf(x_{1:\ell-1})$ such that $q(\by|\bx) :=
q(\by|x_\ell,\bz_\ell)$. This can be viewed as the process $\bX
\rightarrow(\bZ_\ell, X_\ell) \rightarrow \bY$. According to the data processing
inequality, 
\begin{equation}
    I^{q}(X_{1:\ell};Y_{1:L-\ell}) \leq I^{q}(\boldsymbol{Z}_{\ell}, X_{\ell};Y_{1:L-\ell}) \leq \mathcal{H}^{q}(\boldsymbol{Z}_{\ell}, X_{\ell})\leq \mathcal{H}^{q}(\boldsymbol{Z}_{\ell}) + H^{q}( X_{\ell}),
\end{equation}
where we use $\mathcal{H}$ to denote a generalized notion of entropy which
measures the amount of information that can be stored in $\boldsymbol{Z}_\ell$.
We only care about the scaling of $\mathcal{H}$, so its exact definition is
irrelevant to our discussion.

Under the AOD assumption, neural networks store distinct information in almost
orthogonal directions of the hidden state. Therefore, $\mathcal{H}$ should scale
at most logarithmically with respect to the number of AODs as the state size
increases. According to the Kabatjanskii--Levenstein bound
\cite{KabLev78,Cohn_2014}, given an error tolerance $\varepsilon$, the number of
AODs is upper bounded by $\exp(f(\varepsilon) \cdot \dim(\bz_\ell))$ for some
function $f$ that depends purely on the error threshold. Therefore, the
generalized entropy scales as $\mathcal{H}^{q}(\boldsymbol{Z}_{\ell})
\precsim\log \exp(f(\varepsilon) \cdot \dim(\bz_\ell)) \sim \dim(\bz_\ell)$.
Since $H^{q}( X_{\ell}) \le \log(M)$ where $M$ is the vocabulary size, we
conclude 
\begin{equation}
I_{\ell;L}^{\mathrm{BP},q} \leq C \cdot \dim(\boldsymbol{z}_{\ell}) + \log(M).
\end{equation}
\end{proof}

\subsection{Proof Under Lipschitz Continuity Assumption}

The theorem can also be proved assuming only certain Lipschitz continuity
conditions on the neural network.

\begin{theorem}[Theorem~\ref{thm:mi_to_z}, Lipschitz Version]
Assume the history state mapping $\boldf: x_{1:\ell-1} \mapsto \bz_{\ell}$
satisfies $\norm{\boldf(x_{1:{\ell-1}}) - \boldf(x'_{1:{\ell-1}})}_2 \le K_f
\mathbbm{1}(x_{1:{\ell-1}} \ne x'_{1:{\ell-1}})$ and the neural network is
entropy-Lipschitz, satisfying $\abs{H^q(\bY|\bz_{\ell}) - H^q(\bY|\bz'_{\ell})}
\le K_{H}\norm{\bz_{\ell} - \bz'_{\ell}}_2$. Then a model's capacity to capture
bipartite mutual information is bounded as
\begin{equation}
I_{\ell;L}^{\mathrm{BP},q} \leq C \cdot \dim(\boldsymbol{z}_{\ell}) + \log(M)
\end{equation}
where $C$ is some constant and $M$ denotes the vocabulary size.
\end{theorem}

\begin{proof}
We start with the data processing inequality and rewrite the bound as
\begin{equation}
\begin{aligned}
    I^{q}(X_{1:\ell};Y_{1:L-\ell}) &\leq I^{q}(\boldsymbol{Z}_{\ell}, X_{\ell};Y_{1:L-\ell}) \\
    &= I^{q}(\boldsymbol{Z}_{\ell};Y_{1:L-\ell}) + I^{q}(X_{\ell};Y_{1:L-\ell}|\bZ_\ell) \\
    &\le I^{q}(\boldsymbol{Z}_{\ell};Y_{1:L-\ell}) + \log(M),
\end{aligned}
\end{equation}    
where the last inequality uses $I^{q}(X_{\ell};Y_{1:L-\ell}|\bZ_\ell) \le
H^{q}(X_{\ell}) \le \log(M)$, with $M$ being the vocabulary size. 

The history state is a function of the input tokens $\bz_\ell =
\boldf(x_{1:{\ell-1}})$, with $x_{1:{\ell-1}} \in \{1,2,\dots,M\}^{\ell-1}$.
Under our assumption on $\boldf$, $\bz_{\ell}$ lives in a $d$-dimensional ball
of radius $K_f$, where $d = \dim(\bz_{\ell})$.

Consider a quantization $\bQ(\bz_{\ell})$ that maps each $\bz_{\ell}$ to the
nearest point in an $\varepsilon$-covering of this ball. Then
\begin{equation}
    I^{q}(\bZ_{\ell};\bY) = I^{q}(\bQ(\bZ_{\ell});\bY) + H^q(\bY|\bQ(\bZ_{\ell})) - H^q(\bY|\bZ_{\ell}).
\end{equation}

By the entropy-Lipschitz assumption, $H^q(\bY|\bQ(\bZ_{\ell})) -
H^q(\bY|\bZ_{\ell}) \le K_H\varepsilon$. Since $\bQ(\bZ_{\ell})$ is discrete and
takes at most $(2 K_f / \varepsilon)^d$ values (by a covering number argument),
we have $I^{q}(\bQ(\bZ_{\ell});\bY) \le H^q(\bQ(\bZ_{\ell})) \le d \log (2 K_f /
\varepsilon)$. 

Therefore, $I^{q}(\bZ_{\ell};\bY) \le d \log (2 K_f / \varepsilon) +
K_H\varepsilon \le C \cdot d$ for some constant $C$ (by choosing $\varepsilon$
appropriately). This concludes
\begin{equation}
    I_{\ell;L}^{\mathrm{BP},q} \leq C \cdot d + \log(M) = C \cdot \dim(\boldsymbol{z}_{\ell}) + \log(M).
\end{equation}
\end{proof}

\subsection{Discussion}

We believe this theorem is more universal and can be proved in additional ways,
such as by connecting it to channel capacity and potentially showing
$I_{\ell;L}^{\mathrm{BP},q} \leq d \log(1 + \mathrm{SNR}) + \log(M)$. We also
believe the theorem can be established with more relaxed assumptions, similar to
how information dimension is proved to be the upper bound of lossless
compression of continuous random variables \cite{333868}. However, additional
proofs are beyond the scope of this work, and the three proofs provided should
already be applicable in any practical settings.

\section{Additional Details on Experimental Setup} \label{app:experiment}

In this section, we provide detailed information about our experimental setup,
including dataset construction, model configurations, training procedures, and
evaluation metrics.

\subsection{Synthetic Gaussian Distribution Dataset}
\label{app:experiment_gaussian}

For experiments on the multivariate Gaussian distribution, we use the sub-volume
Gaussian distributions described in Appx.~\ref{app:gaussian}, which exhibits
power-law bipartite mutual information scaling with an exponent of $0.79$. To
fully stress the LLMs, we stack 64 copies of the distribution and group the 64
Gaussian variables at each position to form a single token. More specifically,
an example sample looks like
\begin{equation}
\begin{aligned}
\bW &= (W_1,W_2,\dots,W_L)\\
&:=((Z_{1,1}, Z_{1,2},\dots,Z_{1,64}), (Z_{2,1}, Z_{2,2},\dots,Z_{2,64}),\dots,(Z_{L,1}, Z_{L,2},\dots,Z_{L,64})),
\end{aligned}
\end{equation}
where the two subscripts $(i,j)$ refer to the $i$th random variable from the
$j$th copy. In this way, the bipartite mutual information matches better with
natural language, not only in scaling, but also in magnitude (multiplicative
constant). We additionally prepend an all-zero token $W_0$ to each sample to
mimic the effect of the BOS token.

In order to process continuous random variables, we replace the embedding layers
of GPT2 and Mamba(2) models with two-layer MLPs. For output, since all the
conditional distributions are also Gaussian, we use a different two-layer MLP to
output the 64 conditional means ${\mu_{q_{Z_{i,j}|Z_{0:i-1,j}}}}$ and standard
deviations ${\sigma_{q_{Z_{i,j}|Z_{0:i-1,j}}}}$.

As discussed in Appx.~\ref{app:syn_dist_prop}, due to the analytical
construction, the Gaussian distribution permits efficient calculation of
conditional probabilities. Therefore, instead of simply training the neural
networks with negative log likelihood on samples alone, we use the average
conditional KL-divergence estimated as
\begin{equation}
\begin{aligned}
&D_{KL}(p||q_{\theta}) = \\
&\expect_{p_{\bZ}}\left[ \frac{1}{L}\sum_{i=1}^L \frac{1}{64}\sum_{j=1}^{64} \left(\log \frac{\sigma_{q_{Z_{i,j}|Z_{0:i-1,j}}}}{\sigma_{p_{Z_{i,j}|Z_{0:i-1,j}}}} + \frac{\sigma_{p_{Z_{i,j}|Z_{0:i-1,j}}}^2 + (\mu_{q_{Z_{i,j}|Z_{0:i-1,j}}} - \mu_{p_{Z_{i,j}|Z_{0:i-1,j}}})^2}{2 \sigma_{q_{Z_{i,j}|Z_{0:i-1,j}}}^2} - \frac{1}{2}\right)\right]
\end{aligned}
\end{equation}
to reduce sampling variance.

\subsection{Natural Language Dataset (\PG{})}

For the \PG{} dataset, we train on standard average negative log likelihood. We
first split the dataset into samples with a length of approximately 1.2 times
the target length, ensuring each sample starts at the beginning of a sentence.
We then train the models for 5 epochs (approximately 450k iterations) with a
batch size of 16,384 tokens. To maintain consistency across different models, we
always use the same tokenizer from GPT-Neo-X
\cite{black2022gptneox20bopensourceautoregressivelanguage}.

\subsection{Training Configuration}

For the Gaussian distribution training, during each iteration, we use a batch
size of 4 (4 times sequence length number of tokens) with freshly generated
samples, meaning we never reuse any sample. We therefore have effectively a
single epoch, thanks to the ``infinite'' dataset size. We train all neural
networks using the AdamW optimizer and a cosine decay scheduler with warmup. We
use a peak learning rate of $5 \times 10^{-5}$, a weight decay of 0.01, 2000
warmup steps, and 500,000 training steps in total. The results reported are at
the end of training.

For the \PG{} dataset experiments, we use similar hyperparameters: AdamW
optimizer with a cosine decay scheduler with warmup, peak learning rate of $5
\times 10^{-5}$, weight decay of 0.01, 2000 warmup steps, and 500,000 steps in
total. The results reported are at the end of training using a separate
evaluation dataset containing 10,000 samples.

\subsection{Evaluation Metrics} \label{app:experiment_eval_metric}

In this paper, we report results on the position-wise conditional KL-divergence
\begin{equation}
    D_{KL,i} = D_{KL}(p_{W_i|W_{1:i-1}}||q_{W_i|W_{1:i-1}}) = \expect_{p} \left[\log p(W_i|W_{1:i-1}) - \log q(W_i|W_{1:i-1})\right],
\end{equation}
average KL-divergence
\begin{equation}
    D_{KL}^{\mathrm{avg}} = \frac{1}{L} \sum_{i=1}^L D_{KL,i},
\end{equation}
and position-wise conditional NLL
\begin{equation}
    \mathrm{NLL}_i = - \expect_{p} \left[ \log q(W_i|W_{1:i-1})\right].
\end{equation}
One can also define an average NLL as
\begin{equation}
    \mathrm{NLL}^{\mathrm{avg}} = \frac{1}{L} \sum_{i=1}^L \mathrm{NLL}_i,
\end{equation}
which we use in Appx.~\ref{app:add_experiment}.

\subsubsection{Understanding the Behavior of KL-divergence and NLL}
\label{app:nll_explanation}

It is important to understand how KL-divergence and NLL behave differently as
token position increases. Using the relationship between cross-entropy and
KL-divergence from Eq.~\eqref{eq:nll_kl_entropy}, we can decompose the
conditional NLL as
\begin{equation}
    \mathrm{NLL}_i = D_{KL,i} + H^p(W_i|W_{1:i-1}),
\end{equation}
where $H^p(W_i|W_{1:i-1}) = -\expect_{p}[\log p(W_i|W_{1:i-1})]$ is the
conditional entropy of the true distribution at position $i$.

This decomposition reveals why KL-divergence and NLL exhibit opposite trends
with token position. As token position increases, the conditional entropy
$H^p(W_i|W_{1:i-1})$ typically decreases because more context is available,
making the next token more predictable. In natural language, this reflects that
with more preceding text, there is less uncertainty about what comes next.
Meanwhile, the conditional KL-divergence $D_{KL,i}$ often increases with
position, because learning all long-range dependencies becomes more challenging,
resulting in worse model estimations compared to the true conditional
distribution.

For models with sufficient capacity relative to sequence length, the decrease in
conditional entropy dominates, causing NLL to decrease with position despite
increasing KL-divergence. However, for models with insufficient capacity (such
as fixed-state models at long sequence lengths), the KL-divergence can increase
rapidly enough that NLL plateaus or even increases at later positions. This
behavior is precisely what we observe in our experiments with Mamba models in
Fig.~\ref{fig:real_nll}, where Mamba's NLL plateaus at later positions while
transformers' NLL continues to improve.

The same reasoning applies to average quantities: $\mathrm{NLL}^{\mathrm{avg}}$
typically decreases with sequence length $L$ as the average conditional entropy
decreases, while $D_{KL}^{\mathrm{avg}}$ may increase if the model's capacity
becomes insufficient relative to the growing sequence length.

When reporting conditional NLL, we smooth the curves using a small window around
nearby tokens to reduce noise in the results.

\subsection{Model Configurations}

In Tables~\ref{tab:model_config_gaussian} and~\ref{tab:model_config_pg19}, we
include the model configurations and sequence lengths for all experiments
performed in this paper.

\begin{table}[htbp]
    \centering
    \caption{Models and configurations for synthetic dataset experiments.}
    \begin{tabular}{lccc}
        \toprule
        \textbf{Model} & \textbf{num\_hidden\_layers} & \textbf{hidden\_size} &
        \textbf{seq\_len} \\
        \midrule
        GPT2 & 12 & 768 & 256,1024,4096 \\
        GPT2-medium & 24 & 1024 & 256,1024,4096 \\
        GPT2-large & 36 & 1280 & 256,1024,4096 \\
        Mamba-50m & 12 & 512 & 64,256,1024,4096,16384 \\
        Mamba-70m & 24 & 512 & 64,256,1024,4096,16384 \\
        Mamba-130m & 24 & 768 & 64,256,1024,4096,16384 \\
        Mamba-370m & 48 & 1024 & 64,256,1024,4096 \\
        Mamba-790m & 48 & 1536 & 64,256,1024,4096 \\
        Mamba-1.4b & 48 & 2048 & 64,256,1024,4096 \\
        Mamba2-130m & 24 & 768 & 256,1024,4096 \\
        Mamba2-370m & 48 & 1024 & 256,1024,4096 \\
        Mamba2-790m & 48 & 1536 & 256,1024,4096 \\
        \bottomrule
    \end{tabular}
    \label{tab:model_config_gaussian}
\end{table}

\begin{table}[htbp]
    \centering
    \caption{Models and configurations for \PG{} experiments.}
    \begin{tabular}{lccc}
        \toprule
        \textbf{Model} & \textbf{num\_hidden\_layers} & \textbf{hidden\_size} &
        \textbf{seq\_len} \\
        \midrule
        GPT2 & 12 & 768 & 4096,8192 \\
        GPT2-medium & 24 & 1024 & 4096,8192 \\
        GPT2-large & 36 & 1280 & 4096,8192 \\
        Mamba-130m & 24 & 768 & 4096,8192 \\
        Mamba-370m & 48 & 1024 & 4096,8192 \\
        Mamba-790m & 48 & 1536 & 4096,8192 \\
        \bottomrule
    \end{tabular}
    \label{tab:model_config_pg19}
\end{table}

\subsection{Computational Resources and Implementation Details}

Our experiments are performed primarily on H100 GPUs, with varying VRAM sizes
between 80GB and 96GB. Some experiments use A100 GPUs with 80GB VRAM instead. We
use the \texttt{vLLM} library \cite{kwon2023efficientmemorymanagementlarge} when
running inference to estimate the mutual information scaling. For both LLaMA 3.1
405B and DeepSeek V3, we run the FP8 version using 8 H100 GPUs (with 96GB VRAM
each). The model weights and configurations are downloaded from HuggingFace
\cite{wolf2020huggingfacestransformersstateoftheartnatural}.

When training GPT and Mamba(2) models on the Gaussian distribution, we use our
custom library developed in \texttt{PyTorch}
\cite{paszke2019pytorchimperativestylehighperformance}. When training GPT and
Mamba models on the \PG{} dataset, we use the trainer from the HuggingFace
\texttt{transformers} library. All models are initialized from scratch, with
model configurations taken from HuggingFace. All training experiments are
performed on individual H100 and A100 GPUs with FP32 precision to avoid possible
training failures. Although training with FP16 would make the experiments run
faster, it should not affect the actual results. We note that for Mamba2, we use
the official implementation instead of the HuggingFace version. For the GPT2
experiments on \PG{}, we re-implement the attention mechanism with FlexAttention
\cite{li2024flexattentionefficienthighresolutionvisionlanguage} to save memory,
as the official FlashAttention
\cite{dao2022flashattentionfastmemoryefficientexact,dao2023flashattention2fasterattentionbetter,shah2024flashattention3fastaccurateattention}
does not support FP32 precision.

\subsection{Code Availability}

The code for reproducing our mutual information estimation and the \PG{} results
is available at \url{https://github.com/LSquaredM/mutual_info_scaling_law}.

\section{Additional Experimental Results} \label{app:add_experiment}
\begin{figure}[h!]
    \centering
    \includegraphics[width=0.58\linewidth]{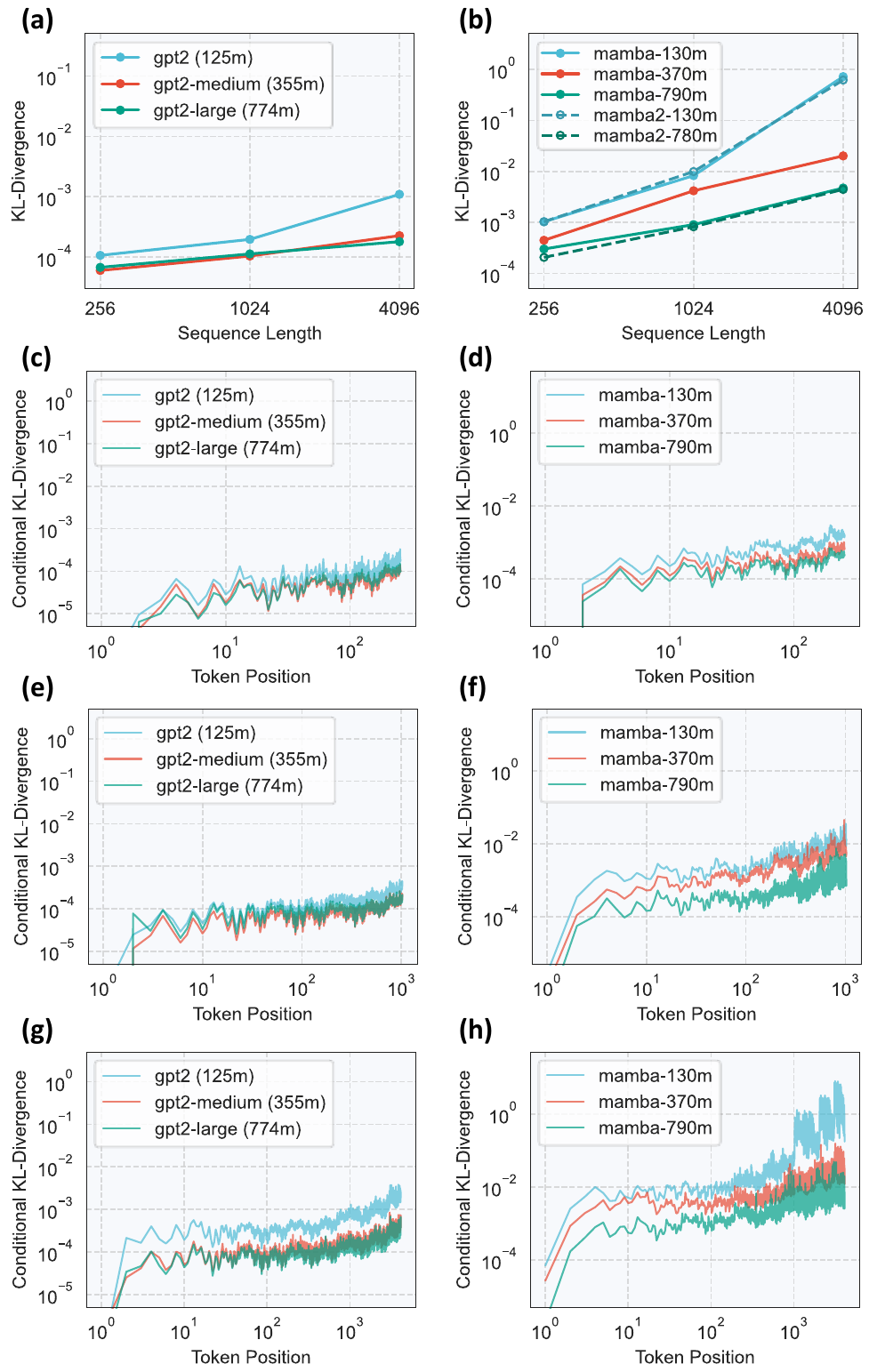}
    \caption{Evaluation of KL-divergence across model architectures trained on
    sub-volume Gaussian distributions. (a, b) Average KL-divergence per token
    for models trained on different sequence lengths [same as
    Fig.~\ref{fig:syn_kl_div} (a, b)]. (c, d) Position-wise conditional
    KL-divergence for models trained on sequence length 256. (e, f)
    Position-wise conditional KL-divergence for models trained on sequence
    length 1024. (g, h) Position-wise conditional KL-divergence for models
    trained on sequence length 4096 [same as Fig.~\ref{fig:syn_kl_div} (c, d)].
    Lower values indicate better performance.}
    \label{fig:gaussian_many}
\end{figure}

In this section, we show additional experimental results. In
Fig.~\ref{fig:gaussian_many}, we include positional-wise conditional
KL-divergences of models trained on sub-volume Gaussian distributions with
sequence length 256 (c, d), 1024 (e, f), and 4096 (g, h). As clearly
demonstrated in the figure, for short sequence lengths, Mamba maintains similar
performances to GPT2; Mamba models of different sizes also appear to have a
smaller performance gap. However, as we go to longer sequence lengths, smaller
Mamba models starts to fail, while GPT2 always maintain relatively stable
performances, consistent with our theory.

\begin{figure}[h!]
    \centering
    \includegraphics[width=0.58\linewidth]{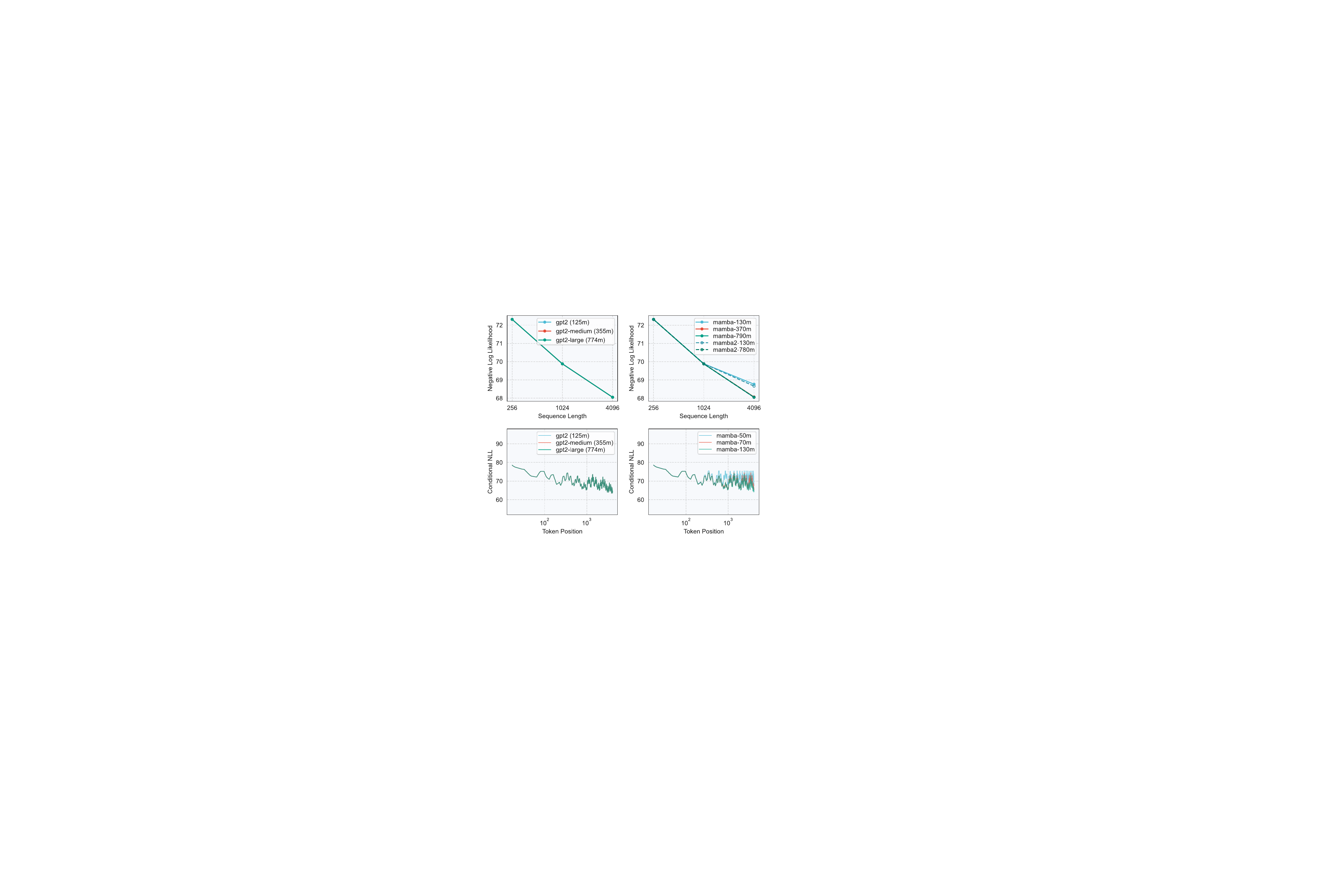}
    \caption{Negative log likelihood (NLL) across model architectures trained on sub-volume Gaussian distributions (a, b) Average NLL per token for models trained on different sequence lengths. (c, d) Position-wise conditional NLL for models trained on sequence length 4096. Lower values indicate better performance.}
    \label{fig:gaussian_nll}
\end{figure}

In Fig.~\ref{fig:gaussian_nll}, we show the negative log likelihood (NLL) of
models trained on sub-volume Gaussian distributions. We note that, because NLL
combines the KL-divergence with the intrinsic entropy of the underlying
distribution (the average and position-wise conditional of which decays as
sequence lengths), the differences between model performances are less visible.
It's worth noting that, since Gaussian random variables are continuous, NLL
values can differ by an arbitrary additive constant by rescaling the
distribution. Therefore, the exact values of conditional NLL do not carry
intrinsic meaning, though relative comparisons (which is exactly the same as the
KL-divergence) between models remain valid.

\begin{figure}[t]
    \centering
    \includegraphics[width=\linewidth]{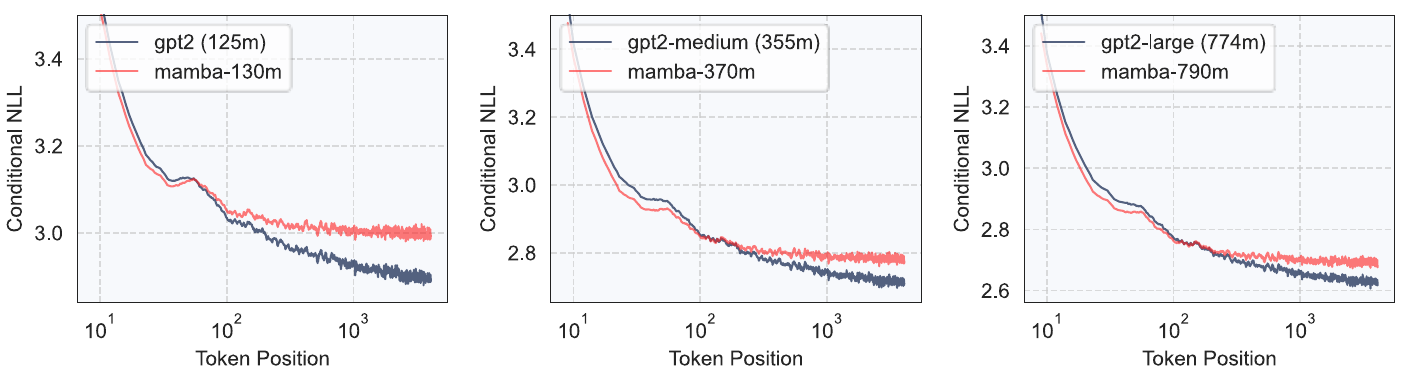}
    \caption{Position-wise conditional negative log likelihood (NLL) evaluation
    for models trained on 4096-token sequences on the \PG{} dataset
    \cite{pg19}.}
    \label{fig:real_nll_4096}
\end{figure}

In Fig.~\ref{fig:real_nll_4096}, we show the position-wise conditional negative
log likelihood (NLL) of models trained on the \PG{} dataset \cite{pg19} with
4096-token sequences. The results here is consistent with the
8192-token-sequence results in the main text.

\end{document}